\documentclass[11pt,a4paper]{article}
\usepackage[hyperref]{naaclhlt2018}
\usepackage{times}
\usepackage{latexsym}
\usepackage[inline]{enumitem}
\usepackage{amsfonts, amsmath}
\usepackage{graphicx}  %Required
\usepackage{multirow}
\usepackage{boldline}
\usepackage{cleveref}

\usepackage{url}

\aclfinalcopy % Uncomment this line for the final submission
\title{Distributional Inclusion Vector Embedding \\ for Unsupervised Hypernymy Detection}

\author{Haw-Shiuan Chang$^1$, ZiYun Wang$^2$, Luke Vilnis$^1$, Andrew McCallum$^1$\\
$^1$University of Massachusetts, Amherst, USA \\
$^2$Tsinghua University, Beijing, China \\
\texttt{hschang@cs.umass.edu, wang-zy14@mails.tsinghua.edu.cn,} \\
\texttt{\{luke, mccallum\}@cs.umass.edu} }

\date{}

\newcommand{\smallt}[1]{{\small{\sf #1}}}
\setlist[itemize]{leftmargin=*}

\crefformat{footnote}{#2\footnotemark[#1]#3}

\begin{document}
\maketitle
\begin{abstract}
Modeling hypernymy, such as \smallt{poodle is-a dog}, is an important generalization aid to many NLP tasks, such as entailment, coreference, relation extraction, and question answering. Supervised learning from labeled hypernym sources, such as WordNet, limits the coverage of these models, which can be addressed by learning hypernyms from unlabeled text.  Existing unsupervised methods either do not scale to large vocabularies or yield unacceptably poor accuracy.  This paper introduces {\it distributional inclusion vector embedding (DIVE)}, a simple-to-implement unsupervised method of hypernym discovery via per-word non-negative vector embeddings which preserve the inclusion property of word contexts in a low-dimensional and interpretable space. In experimental evaluations more comprehensive than any previous literature of which we are aware---evaluating on 11 datasets using multiple existing as well as newly proposed scoring functions---we find that our method provides up to double the precision of previous unsupervised embeddings, and the highest average performance, using a much more compact word representation, and yielding many new state-of-the-art results.
\end{abstract}

\section{Introduction}

Numerous applications benefit from compactly representing context distributions, which assign meaning to objects under the rubric of \emph{distributional semantics}. In natural language processing, distributional semantics has long been used to assign meanings to words (that is, to \emph{lexemes} in the dictionary, not individual instances of word tokens). The meaning of a word in the distributional sense is often taken to be the set of textual contexts (nearby tokens) in which that word appears, represented as a large sparse bag of words (SBOW). Without any supervision, Word2Vec~\citep{mikolov2013distributed}, among other approaches based on matrix factorization \citep{levy2015improving}, successfully compress the SBOW into a much lower dimensional embedding space, increasing the scalability and applicability of the embeddings while preserving (or even improving) the correlation of geometric embedding similarities with human word similarity judgments. 

While embedding models have achieved impressive results, context distributions capture more semantic information than just word similarity. The \emph{distributional inclusion hypothesis} (DIH) \citep{weeds2003general,geffet2005distributional,cimiano2005learning} posits that the context set of a word tends to be a subset of the contexts of its hypernyms. For a concrete example, most adjectives that can be applied to \smallt{poodle} can also be applied to \smallt{dog},  because \smallt{dog} is a hypernym of \smallt{poodle} (e.g. both can be \smallt{obedient}). However, the converse is not necessarily true --- a \smallt{dog} can be \smallt{straight-haired} but a \smallt{poodle} cannot. Therefore, \smallt{dog} tends to have a broader context set than \smallt{poodle}. Many asymmetric scoring functions comparing SBOW features based on DIH have been developed for hypernymy detection \citep{weeds2003general,geffet2005distributional,SantusSS17}.

Hypernymy detection plays a key role in many challenging NLP tasks, such as textual entailment~\citep{sammons2011recognizing}, coreference~\citep{ponzetto2006exploiting}, relation extraction~\citep{DemeesterRR16} and question answering~\citep{huang2008question}. Leveraging the variety of contexts and inclusion properties in context distributions can greatly increase the ability to discover taxonomic structure among words~\citep{SantusSS17}. The inability to preserve these features limits the semantic representation power and downstream applicability of some popular unsupervised learning approaches such as Word2Vec.
%summarization~\citep{pasunuru2017towards},

Several recently proposed methods aim to encode hypernym relations between words in dense embeddings, such as Gaussian embedding~\citep{VilnisM14,athiwaratkun2017multimodal}, Boolean Distributional Semantic Model~\citep{KruszewskiPB15}, order embedding~\citep{VendrovKFU15}, H-feature detector~\citep{RollerE16}, HyperVec~\citep{NguyenKWV17}, dual tensor~\citep{glavavs2017dual}, Poincar{\'{e}} embedding \citep{NickelK17}, and LEAR~\citep{vulic2017specialising}. However, the methods focus on supervised or semi-supervised settings where a massive amount of hypernym annotations are available~\citep{VendrovKFU15,RollerE16,NguyenKWV17,glavavs2017dual,vulic2017specialising}, do not learn from raw text \citep{NickelK17} or lack comprehensive experiments on the hypernym detection task~\citep{VilnisM14,athiwaratkun2017multimodal}.

Recent studies~\citep{levy2015supervised,SantusSS17} have underscored the difficulty of generalizing supervised hypernymy annotations to unseen pairs --- classifiers often effectively memorize prototypical hypernyms (`general' words) and ignore relations between words. These findings motivate us to develop more accurate and scalable unsupervised embeddings to detect hypernymy and propose several scoring functions to analyze the embeddings from different perspectives.
%Additionally, there is a lack of annotated hypernymy datasets for domains such as scientific literature. Although~\citet{RollerE16} demonstrate that the hypernym relations between words can be learned by carefully avoiding lexicon memorization, our experiment results show that unsupervised 
%This practical need motivates the development of more accurate and scalable unsupervised approaches to hypernymy detection.

\subsection{Contributions}
\begin{itemize}
\itemsep0em 
\item A novel unsupervised low-dimensional embedding method via performing non-negative matrix factorization (NMF) on a weighted PMI matrix, which can be efficiently optimized using modified skip-grams. 
\item Theoretical and qualitative analysis illustrate that the proposed embedding can intuitively and interpretably preserve inclusion relations among word contexts.
\item Extensive experiments on 11 hypernym detection datasets demonstrate that the learned embeddings dominate previous low-dimensional unsupervised embedding approaches, achieving similar or better performance than SBOW, on both existing and newly proposed asymmetric scoring functions, while requiring much less memory and compute.
%\item A novel unsupervised low-dimensional embedding method to model inclusion relations among word contexts via performing non-negative matrix factorization (NMF) on a weighted PMI matrix, which can be efficiently optimized using modified skip-grams. 
%\item Several new asymmetric comparison functions to measure inclusion and generality properties and to evaluate different aspects of unsupervised embeddings.
%\item Extensive experiments on 11 datasets demonstrate the learned embeddings and comparison functions achieve state-of-the-art performances on unsupervised hypernym detection while requiring much less memory and compute than approaches based on the full SBOW. 
\end{itemize}

\section{Method}

The \emph{distributional inclusion hypothesis} (DIH) suggests that the context set of a hypernym tends to contain the context set of its hyponyms. When representing a word as the counts of contextual co-occurrences, the count in every dimension of hypernym $y$ tends to be larger than or equal to the corresponding count of its hyponym $x$:
\begin{equation}\label{eq-DIH}
x\preceq y \iff \forall c \in V, \; \#(x,c) \leq \#(y,c), 
\end{equation}
where $x \preceq y$ means $y$ is a hypernym of $x$, $V$ is the set of vocabulary, and $\#(x,c)$ indicates the number of times that word $x$ and its context word $c$ co-occur in a small window with size $|W|$ in the corpus of interest $D$. Notice that the concept of DIH could be applied to different context word representations. For example, \citet{geffet2005distributional} represent each word by the set of its co-occurred context words while discarding their counts. In this study, we define the inclusion property based on counts of context words in~\eqref{eq-DIH} because the counts are an effective and noise-robust feature for the hypernymy detection using only the context distribution of words~\citep{clarke2009context,VulicGKHK16,SantusSS17}. 
%discard the counts of context words and only consider whether two words co-occur at least once in the corpus instead

Our goal is to produce lower-dimensional embeddings preserving the inclusion property that the embedding of hypernym $y$ is larger than or equal to the embedding of its hyponym $x$ in every dimension. Formally, the desired property can be written as 
\begin{equation}\label{eq-reversed-order}
  x\preceq y \iff \mathbf{x}[i] \leq \mathbf{y}[i] \; ,\forall i \in \{1,...,L \},
\end{equation}
where $L$ is number of dimensions in the embedding space. We add additional non-negativity constraints, i.e. $\mathbf{x}[i]\ge 0,\mathbf{y}[i] \ge 0, \forall i$, in order to increase the interpretability of the embeddings (the reason will be explained later in this section).
%i.e. $x[i]\ge 0,y[i] \ge 0, \forall i$, the utility of which is explained later in this section.

This is a challenging task. In reality, there are a lot of noise and systematic biases that cause the violation of DIH in Equation~\eqref{eq-DIH} (i.e. $\#(x,c)>\#(y,c)$ for some neighboring word $c$), but the general trend can be discovered by processing thousands of neighboring words in SBOW together~\citep{SantusSS17}. After the compression, the same trend has to be estimated in a much smaller embedding space which discards most of the information in SBOW, so it is not surprising to see most of the unsupervised hypernymy detection studies focus on SBOW~\citep{SantusSS17} and the existing unsupervised embedding methods like Gaussian embedding have degraded accuracy~\citep{VulicGKHK16}.
 %In SBOW, it is common that a target word has several thousand neighboring words.

%levy2014neural,levy2015improving
\subsection{Inclusion Preserving Matrix Factorization}
Popular methods of unsupervised word embedding are usually based on matrix factorization~\citep{levy2015improving}. The approaches first compute a co-occurrence statistic between the $w$th word and the $c$th context word as the $(w,c)$th element of the matrix $M[w,c]$. Next, the matrix $M$ is factorized such that $M[w,c] \approx \mathbf{w}^T \mathbf{c}$, where $\mathbf{w}$ is the low dimension embedding of $w$th word and $\mathbf{c}$ is the $c$th context embedding. 

The statistic in $M[w,c]$ is usually related to pointwise mutual information~\citep{levy2015improving}: $PMI(w,c)=\log(\frac{P(w,c)}{P(w)\cdot P(c)})$, where $P(w,c)=\frac{\#(w,c)}{|D|}$, $|D|=\sum\limits_{w \in V} \sum\limits_{c \in V} \#(w,c)$ is number of co-occurrence word pairs in the corpus, $P(w)=\frac{\#(w)}{|D|}$, $\#(w) = \sum\limits_{c \in V} \#(w,c)$ is the frequency of the word $w$ times the window size $|W|$, and similarly for $P(c)$. For example, $M[w,c]$ could be set as positive PMI (PPMI), $\max(PMI(w,c),0)$, or shifted PMI, $PMI(w,c)-\log(k')$, which \citep{levy2014neural} demonstrate is connected to skip-grams with negative sampling (SGNS). 

Intuitively, since $M[w,c] \approx \mathbf{w}^T \mathbf{c}$, larger embedding values of $\mathbf{w}$ at every dimension seems to imply larger $\mathbf{w}^T \mathbf{c}$, larger $M[w,c]$, larger $PMI(w,c)$, and thus larger co-occurrence count $\#(w,c)$. However, the derivation has two flaws: 
\begin{enumerate*}[label=(\arabic*)]
	\item $\mathbf{c}$ could contain negative values and 
    \item lower $\#(w,c)$ could still lead to larger $PMI(w,c)$ as long as the $\#(w)$ is small enough. %larger $PMI(w,c)$ might come from lower $\#(w,c)$ as long as the $\#(w)$ is small enough.
\end{enumerate*}

To preserve DIH, we propose a novel word embedding method, \emph{distributional inclusion vector embedding (DIVE)}, which fixes the two flaws by performing non-negative factorization (NMF)~\citep{lee2001algorithms} on the matrix $M$, where $M[w,c] = $ \vspace{-3mm}
\begin{equation}
\label{eq:PMI_NW}
\log(\frac{P(w,c)}{P(w)\cdot P(c)}\cdot\frac{\#(w)}{k_I\cdot Z}) = \log(\frac{\#(w,c)|V|}{\#(c)k_I}), 
\end{equation}
%$M[w,c] = \log(\frac{P(w,c)}{P(w)\cdot P(c)}\cdot\frac{\#(w)}{k\cdot Z}) = \log(\frac{\#(w,c)|V|}{\#(c)k})$, 
where $k_I$ is a constant which shifts PMI value like SGNS, $Z=\frac{|D|}{|V|}$ is the average word frequency, and $|V|$ is the vocabulary size. We call this weighting term $\frac{\#(w)}{Z}$ \emph{inclusion shift}.
%The weights

After applying the non-negativity constraint and inclusion shift, the inclusion property in DIVE (i.e. Equation~\eqref{eq-reversed-order}) implies that Equation~\eqref{eq-DIH} (DIH) holds if the matrix is reconstructed perfectly. The derivation is simple: If the embedding of hypernym $\textbf{y}$ is greater than or equal to the embedding of its hyponym $\textbf{x}$ in every dimension ($\mathbf{x}[i] \leq \mathbf{y}[i] \; ,\forall i$), $\textbf{x}^T\textbf{c} \leq \textbf{y}^T\textbf{c}$ since context vector $\textbf{c}$ is non-negative. Then, $M[x,c] \leq M[y,c]$ tends to be true because $\textbf{w}^T\textbf{c} \approx M[w,c]$. This leads to $\#(x,c) \leq \#(y,c)$ because $M[w,c] = \log(\frac{\#(w,c)|V|}{\#(c)k_I})$ and only $\#(w,c)$ changes with $w$. %This provides an alternative explanation of why DIVE can preserve the inclusion property.

%The design encourages the inclusion property in DIVE (i.e. Equation~\eqref{eq-reversed-order}) to be satisfied because the property implies that Equation~\eqref{eq-DIH} (DIH) holds if the matrix is reconstructed perfectly. 

%\citeauthor{levy2014neural} prove Equation \eqref{eq-sgns} is equivalent to factorizing a shifted PMI matrix $M$, where $M[w,c] = \log(\frac{P(w,c)}{P(w)\cdot P(c)}\cdot\frac{1}{k})$.
%\begin{equation}\label{eq-sgns}
%\end{equation}

%With the non-negativity constraint and weights on negative samples, DIVE is performing non-negative matrix factorization (NMF) on the context matrix $M'$, where 
%\begin{equation}\label{eq:PMI_NW}
%M'[w,c] = \log(\frac{P(w,c)}{P(w)\cdot P(c)}\cdot\frac{\#(w)}{k\cdot Z}) = \log(\frac{\#(w,c)|V|}{\#(c)k}). 
%\end{equation}
%The weights on negative sampling encourages more frequent words to have larger embeddings. This modification allows us to retain the word frequency signal.

\begin{figure*}[t!]
\centering
\includegraphics[width=1\linewidth]{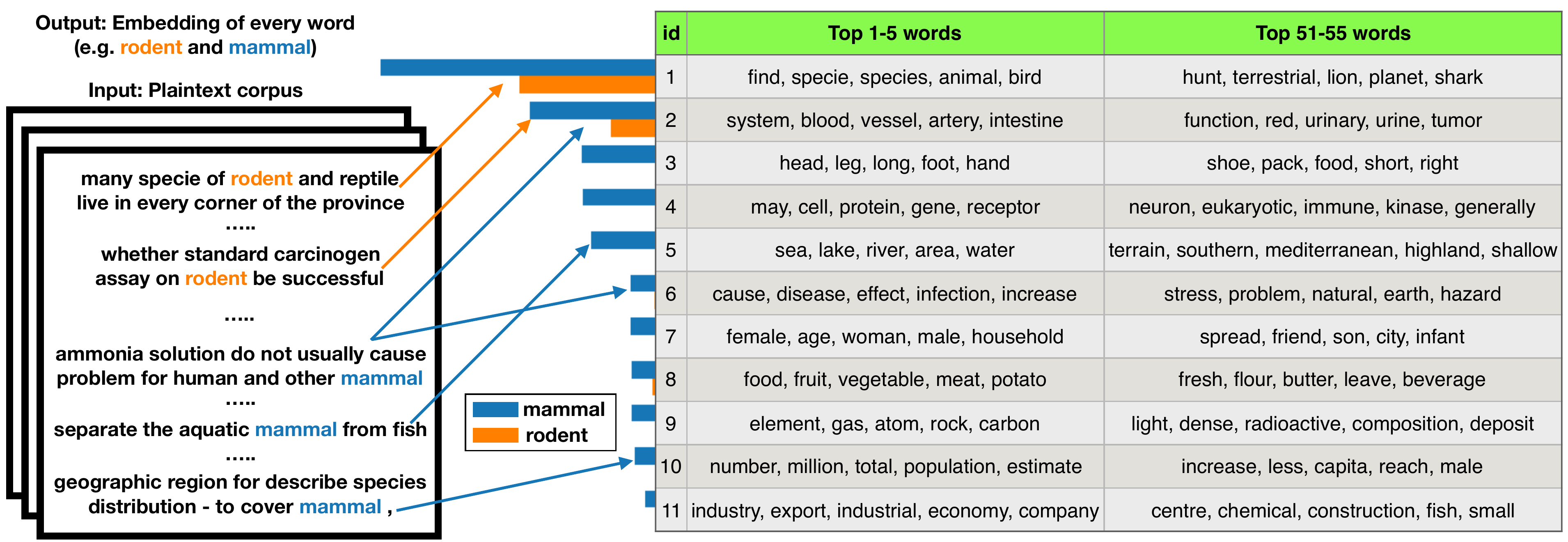}
\caption{The embedding of the words \smallt{rodent} and \smallt{mammal} trained by the co-occurrence statistics of context words using DIVE. The index of dimensions is sorted by the embedding values of \smallt{mammal} and values smaller than 0.1 are neglected. The top 5 words (sorted by its embedding value of the dimension) tend to be more general or more representative on the topic than the top 51-105 words.}
\label{fig:embedding_vis}
\end{figure*}

\subsection{Optimization}
%\subsection{Skip Gram}
Due to its appealing scalability properties during training time~\citep{levy2015improving}, we optimize our embedding based on the skip-gram with negative sampling (SGNS)~\citep{mikolov2013distributed}. The objective function of SGNS is
\begin{equation}\label{eq-SGNS}
\small
\begin{aligned}
&l_{SGNS}=\sum_{w\in V}\sum_{c \in V}\#(w,c)\log\sigma(\mathbf{w}^T\mathbf{c}) ~+ \\
 & \sum_{w\in V} k' \sum_{c \in V}\#(w,c) \underset{c_N\sim P_D}{\mathbb{E}} [\log \sigma(-\mathbf{w}^T\mathbf{c_N})], 
\end{aligned}
\end{equation}
where $\mathbf{w} \in \mathbb{R}$, $\mathbf{c} \in \mathbb{R}$, $\mathbf{c_N} \in \mathbb{R}$, $\sigma$ is the logistic sigmoid function, and $k'$ is a constant hyper-parameter indicating the ratio between positive and negative samples.

\citet{levy2014neural} demonstrate SGNS is equivalent to factorizing a shifted PMI matrix $M'$, where $M'[w,c] = \log(\frac{P(w,c)}{P(w)\cdot P(c)}\cdot\frac{1}{k'})$. By setting $k'=\frac{k_I \cdot Z}{\#(w)}$ and applying non-negativity constraints to the embeddings, DIVE can be optimized using the similar objective function:
\begin{equation}\label{eq-DIVE}
\small
\begin{aligned}
&l_{DIVE}=\sum_{w\in V}\sum_{c \in V}\#(w,c)\log\sigma(\mathbf{w}^T\mathbf{c}) ~+ \\
&k_I\sum_{w\in V} \frac{Z}{\#(w)} \sum_{c \in V}\#(w,c) \underset{c_N\sim P_D}{\mathbb{E}} [\log \sigma(-\mathbf{w}^T\mathbf{c_N})], 
%l=\sum_{w\in V}\sum_{c}\#(w,c)\log\sigma(\mathbf{w}^T\mathbf{c}) ~+ k\sum_{w\in V}\sum_{c}\#(w,c)\underset{c_N\sim P_D}{\mathbb{E}}[\log \sigma(-\mathbf{w}^T\mathbf{c_N})],
%  l=&\sum_{w\in V}\sum_{c}\#(w,c)\log\sigma(\mathbf{w}^T\mathbf{c}) ~+ \\
%    &k\sum_{w\in V}\sum_{c}\#(w,c)\underset{c_N\sim P_D}{\mathbb{E}}[\log \sigma(-\mathbf{w}^T\mathbf{c_N})],
\end{aligned}
\end{equation}
where $\mathbf{w} \geq 0, \mathbf{c} \geq 0, \mathbf{c_N} \geq 0$,  and $k_I$ is a constant hyper-parameter. $P_D$ is the distribution of negative samples, which we set to be the corpus word frequency distribution (not reducing the probability of drawing frequent words like SGNS) in this paper. Equation~\eqref{eq-DIVE} is optimized by ADAM~\citep{kingma2015adam}, a variant of stochastic gradient descent (SGD). The non-negativity constraint is implemented by projection~\citep{polyak1969minimization} (i.e. clipping any embedding which crosses the zero boundary after an update).

The optimization process provides an alternative angle to explain how DIVE preserves DIH. The gradients for the word embedding $\mathbf{w}$ is
\begin{equation}\label{eq-dl}
\small
\begin{aligned}
\frac{dl_{DIVE}}{d\mathbf{w}}= & \sum_{c \in V}\#(w,c)(1-\sigma(\mathbf{w}^T\mathbf{c}))\mathbf{c} ~- \\
& k_I\sum_{c_N \in V} \frac{\#(c_N)}{|V|}\sigma(\mathbf{w}^T\mathbf{c_N})\mathbf{c_N}.
\end{aligned}
\end{equation}
Assume hyponym x and hypernym y satisfy DIH in Equation~\eqref{eq-DIH} and the embeddings $\mathbf{x}$ and $\mathbf{y}$ are the same at some point during the gradient ascent. At this point, the gradients coming from negative sampling (the second term) decrease the same amount of embedding values for both $x$ and $y$. However, the embedding of hypernym $\mathbf{y}$ would get higher or equal positive gradients from the first term than $\mathbf{x}$ in every dimension because $\#(x,c) \leq \#(y,c)$. This means Equation~\eqref{eq-DIH} tends to imply Equation~\eqref{eq-reversed-order} because the hypernym has larger gradients everywhere in the embedding space. 

Combining the analysis from the matrix factorization viewpoint, DIH in Equation~\eqref{eq-DIH} is approximately equivalent to the inclusion property in DIVE (i.e. Equation~\eqref{eq-reversed-order}).
%(assume updates are small enough so none clip to zero)

%\subsection{Non-Negative Skip Gram}
%In our method, we constrain the embedding space to be non-negative for all word vectors $\mathbf{w}$ and context vectors $\mathbf{c}$. Then, whenever a word $w$ has a context word $c$ during the training process, the method will both try to increase the magnitude of the word embedding $\textbf{w}$, and move its vector direction closer to the direction of $\textbf{c}$ in order to maximize $\mathbf{w}^T\mathbf{c}$. This implies that the positive samples of $w$ will encourage $\textbf{w}$ to become a positive weighted summation of the context vectors $\textbf{c}$ of its neighboring word set $\{c|\#(w,c)>0\}$. Since by the DIH, a hypernym tends to contain the contexts of all of its hyponyms, the positive samples in the objective function encourage this distributional inclusion characteristic.

\subsection{PMI Filtering}
%frequency signal is sometimes useful
For a frequent target word, there must be many neighboring words that incidentally appear near the target word without being semantically meaningful, especially when a large context window size is used. The unrelated context words cause noise in both the word vector and the context vector of DIVE. We address this issue by filtering out context words $c$ for each target word $w$ when the PMI of the co-occurring words is too small (i.e. $\log(\frac{P(w,c)}{P(w)\cdot P(c)})<\log(k_f)$). That is, we set $\#(w,c) = 0$ in the objective function. This preprocessing step is similar to computing PPMI in SBOW~\citep{bullinaria2007extracting}, where low PMI co-occurrences are removed from SBOW. 

\subsection{Interpretability}
\label{sec:interp}
After applying the non-negativity constraint, we observe that each latent factor in the embedding is interpretable as previous findings suggest~\citep{pauca2004text,murphy2012learning} (i.e. each dimension roughly corresponds to a topic). Furthermore, DIH suggests that a general word appears in more diverse contexts/topics. By preserving DIH using inclusion shift, the embedding of a general word (i.e. hypernym of many other words) tends to have larger values in these dimensions (topics). This gives rise to a natural and intuitive interpretation of our word embeddings: the word embeddings can be seen as unnormalized probability distributions over topics. In Figure~\ref{fig:embedding_vis}, we visualize the unnormalized topical distribution of two words, \smallt{rodent} and \smallt{mammal}, as an example. Since \smallt{rodent} is a kind of \smallt{mammal}, the embedding (i.e. unnormalized topical distribution) of \smallt{mammal} includes the embedding of \smallt{rodent} when DIH holds. More examples are illustrated in our supplementary materials.

\section{Unsupervised Embedding Comparison}
\label{sec:emb_comparison}
% We evaluate the quality of our embeddings using different hypernym detection scoring functions on both SBOW and DIVE embeddings. 

In this section, we compare DIVE with other unsupervised hypernym detection methods. In this paper, unsupervised approaches refer to the methods that only train on plaintext corpus without using any hypernymy or lexicon annotation.

\begin{table*}[t!]
\centering
\scalebox{0.7}{
\begin{tabular}{|c|c|c|c|c|c|c|c|}
\cline{1-7}
Dataset&BLESS&EVALution&LenciBenotto&Weeds&Medical&LEDS \\ \cline{1-7}
Random & 5.3 & 26.6 & 41.2& 51.4 &8.5&50.5 \\
Word2Vec + C & 9.2 & 25.4& 40.8& 51.6& 11.2&71.8\\
GE + C & 10.5 & 26.7 &43.3 & 52.0& 14.9 & 69.7  \\
GE + KL & 7.6 & 29.6 & 45.1 & 51.3& 15.7 & 64.6 (80\cref{GE_perf}) \\
DIVE + C$\cdot\Delta$S &  \textbf{16.3} &\textbf{33.0} &\textbf{50.4} & \textbf{65.5}& \textbf{19.2} &  \textbf{83.5} \\ \hline
\hline \clineB{6-6}{2}
%\multicolumn{6}{|c|}{} \\ \hline
Dataset&TM14&Kotlerman 2010&HypeNet&WordNet & \multicolumn{1}{V{3} c V{3}}{\textbf{Avg (10 datasets)}}& HyperLex \\ \cline{1-7}
Random & 52.0&30.8&24.5&55.2& \multicolumn{1}{V{3} c V{3}}{23.2}&0\\
Word2Vec + C &  52.1 &\textbf{39.5}& 20.7&\textbf{63.0}&\multicolumn{1}{V{3} c V{3}}{25.3}&16.3\\
GE + C &  53.9 & 36.0& 21.6 & 58.2 & \multicolumn{1}{V{3} c V{3}}{26.1} & 16.4 \\
GE + KL &  52.0 & 39.4&23.7 & 54.4 & \multicolumn{1}{V{3} c V{3}}{25.9} &9.6 (20.6\cref{GE_perf}) \\
%DIVE + C$\cdot\Delta$S & \textbf{25.3} &  \textbf{83.5} &\textbf{57.2} & 36.6  \\ \hline
DIVE + C$\cdot\Delta$S & \textbf{57.2} & 36.6 &\textbf{32.0} & 60.9 & \multicolumn{1}{V{3} c V{3}}{\textbf{32.7}}&\textbf{32.8} \\ \clineB{6-6}{3} \hline
%\hline 
\end{tabular}
}
\caption{Comparison with other unsupervised embedding methods. The scores are AP@all (\%) for the first 10 datasets and Spearman $\rho$ (\%) for HyperLex. Avg (10 datasets) shows the micro-average AP of all datasets except HyperLex. Word2Vec+C scores word pairs using cosine similarity on skip-grams. GE+C and GE+KL compute cosine similarity and negative KL divergence on Gaussian embedding, respectively.}
\label{tb:unsup_embedding}
\end{table*}

\subsection{Experiment Setup}
\label{sec:exp_setup}
The embeddings are tested on 11 datasets. The first 4 datasets come from the recent review of~\citet{SantusSS17}\footnote{\small \url{https://github.com/vered1986/UnsupervisedHypernymy}}: BLESS~\citep{baroni2011we}, EVALution~\citep{santus2015evalution}, Lenci/Benotto~\citep{benotto2015distributional}, and Weeds~\citep{weeds2014learning}. The next 4 datasets are downloaded from the code repository of the H-feature detector~\citep{RollerE16}\footnote{\small \url{https://github.com/stephenroller/emnlp2016/}}: Medical (i.e., Levy 2014)~\citep{LevyDG14}, LEDS (also referred to as ENTAILMENT or Baroni 2012)~\citep{baroni2012entailment}, TM14 (i.e., Turney 2014)~\citep{turney2015experiments}, and Kotlerman 2010~\citep{kotlerman2010directional}. In addition, the performance on the test set of HypeNet~\citep{ShwartzGD16} (using the random train/test split), the test set of WordNet~\citep{VendrovKFU15}, and all pairs in HyperLex~\citep{VulicGKHK16} are also evaluated. 

The F1 and accuracy measurements are sometimes very similar even though the quality of prediction varies, so we adopted average precision, AP@all~\citep{zhu2004recall} (equivalent to the area under the precision-recall curve when the constant interpolation is used), as the main evaluation metric. The HyperLex dataset has a continuous score on each candidate word pair, so we adopt Spearman rank coefficient $\rho$~\citep{fieller1957tests} as suggested by the review study of~\citet{VulicGKHK16}. Any OOV (out-of-vocabulary) word encountered in the testing data is pushed to the bottom of the prediction list (effectively assuming the word pair does not have hypernym relation). 

We trained all methods on the first 51.2 million tokens of WaCkypedia corpus~\citep{baroni2009wacky} because DIH holds more often in this subset (i.e. SBOW works better) compared with that in the whole WaCkypedia corpus. The window size $|W|$ of DIVE and Gaussian embedding are set as 20 (left 10 words and right 10 words). The number of embedding dimensions in DIVE $L$ is set to be 100. The other hyper-parameters of DIVE and Gaussian embedding are determined by the training set of HypeNet. Other experimental details are described in our supplementary materials.
%appendix.

\subsection{Results}
\label{sec:exp1}
If a pair of words has hypernym relation, the words tend to be similar (sharing some context words) and the hypernym should be more general than the hyponym. Section~\ref{sec:interp} has shown that the embedding could be viewed as an unnormalized topic distribution of its context, so the embedding of hypernym should be similar to the embedding of its hyponym but having larger magnitude. As in HyperVec~\citep{NguyenKWV17}, we score the hypernym candidates by multiplying two factors corresponding to these properties. The C$\cdot \Delta$S (i.e. the cosine similarity multiply the difference of summation) scoring function is defined as
\begin{equation}
\small
\label{eq:C_dS}
C \cdot \Delta S(\textbf{w}_q \rightarrow \textbf{w}_p) = \frac{\textbf{w}_q^T \textbf{w}_p}{||\textbf{w}_q||_2\cdot||\textbf{w}_p||_2} \cdot (\|\textbf{w}_p\|_1 -\|\textbf{w}_q\|_1),
\end{equation}
where $\textbf{w}_p$ is the embedding of hypernym and $\textbf{w}_q$ is the embedding of hyponym.

As far as we know, Gaussian embedding (GE)~\citep{VilnisM14} is the state-of-the-art unsupervised embedding method which can capture the asymmetric relations between a hypernym and its hyponyms. Gaussian embedding encodes the context distribution of each word as a multivariate Gaussian distribution, where the embeddings of hypernyms tend to have higher variance and overlap with the embedding of their hyponyms. In Table~\ref{tb:unsup_embedding}, we compare DIVE with Gaussian embedding\footnote{\label{GE_perf} Note that higher AP is reported for some models in previous literature: 80~\citep{VilnisM14} in LEDS, 74.2~\citep{athiwaratkun2017multimodal} in LEDS, and 20.6~\citep{VulicGKHK16} in HyperLex. The difference could be caused by different train/test setup (e.g. How the hyper-parameters are tuned, different training corpus, etc.). However, DIVE beats even these results.} using the code implemented by~\citet{athiwaratkun2017multimodal}\footnote{\url{https://github.com/benathi/word2gm}} and with word cosine similarity using skip-grams. The performances of random scores are also presented for reference. As we can see, DIVE is usually significantly better than other unsupervised embedding.

\section{SBOW Comparison}
Unlike Word2Vec, which only tries to preserve the similarity signal, the goals of DIVE cover preserving the capability of measuring not only the similarity but also whether one context distribution includes the other (inclusion signal) or being more general than the other (generality signal).

In this experiment, we perform a comprehensive comparison between SBOW and DIVE using multiple scoring functions to detect the hypernym relation between words based on different types of signal. The window size $|W|$ of SBOW is also set as 20, and experiment setups are the same as that described in Section~\ref{sec:exp_setup}. Notice that the comparison is inherently unfair because most of the information would be lost during the aggressive compression process of DIVE, and we would like to evaluate how well DIVE can preserve signals of interest using the number of dimensions which is several orders of magnitude less than that of SBOW. 

\subsection{Unsupervised Scoring Functions}
After trying many existing and newly proposed functions which score a pair of words to detect hypernym relation between them, we find that good scoring functions for SBOW are also good scoring functions for DIVE. Thus, in addition to C$\cdot \Delta$S used in Section~\ref{sec:exp1}, we also present 4 other best performing or representative scoring functions in the experiment (see our supplementary materials for more details):
\begin{itemize}
\itemsep0em 
\item \textbf{Inclusion}: CDE~\citep{clarke2009context} computes the summation of element-wise minimum over the magnitude of hyponym embedding (i.e. $\frac{||\min(\textbf{w}_p,\textbf{w}_q)||_1}{||\textbf{w}_q||_1}$). CDE measures the degree of violation of equation (1). Equation (1) holds if and only if CDE is 1. Due to noise in SBOW, CDE is rarely exactly 1, but hypernym pairs usually have higher CDE. Despite its effectiveness, the good performance could mostly come from the magnitude of embeddings/features instead of inclusion properties among context distributions. To measure the inclusion properties between context distributions $\textbf{d}_p$ and $\textbf{d}_q$ ($\textbf{w}_p$ and $\textbf{w}_q$ after normalization, respectively), we use negative asymmetric L1 distance ($-AL_1$)\footnote{The meaning and efficient implementation of $AL_1$ are illustrated in our supplementary materials} as one of our scoring function, where
\begin{equation}
\label{eq:AL1}
\begin{aligned}
%\footnotesize
AL_1  = & \min_a \sum_c w_0 \cdot \max(a \textbf{d}_q[c]-\textbf{d}_p[c],0 )+ \\
& \max(\textbf{d}_p[c] - a \textbf{d}_q[c],0 ), 
\end{aligned} 
\end{equation}
and $w_0$ is a constant hyper-parameter.
\item \textbf{Generality}: When the inclusion property in \eqref{eq-reversed-order} holds, $||\textbf{y}||_1 = \sum_i\textbf{y}[i] \geq \sum_i\textbf{x}[i] = ||\textbf{x}||_1$. Thus, we use summation difference ($||\textbf{w}_p||_1-||\textbf{w}_q||_1$) as our score to measure generality signal ($\Delta$S).
\item \textbf{Similarity plus generality}: Computing cosine similarity on skip-grams (i.e. Word2Vec + C in Table~\ref{tb:unsup_embedding}) is a popular way to measure the similarity of two words, so we multiply the Word2Vec similarity with summation difference of DIVE or SBOW (W$\cdot\Delta$S) as an alternative of C$\cdot \Delta$S.

\end{itemize}
%scoring function to use the signal
%How well we can use the signal to predict hypernym relation between words. 

%In the experiment, we first compare the DIVE with previous state-of-the-art results using C$\cdot\Delta$S and W$\cdot\Delta$S. 
%In order to know how well each component in DIVE preserves signal of interest in SBOW, we perform comprehensive comparison between SBOW and DIVE. 

%\subsection{Compression of SBOW}
%In this experiment, we examine whether DIVE successfully preserves the signals for hypernymy detection tasks, which are measured by the same scoring functions designed for SBOW. Summation difference ($\Delta$S) and CDE perform the best among generality and inclusion functions in Table~\ref{tb:metrics}, respectively. $AL_1$ could be used to examine the inclusion properties after removing the frequency signal. Therefore, we will present the results using these 3 scoring functions, along with W$\cdot\Delta$S and C$\cdot\Delta$S.

\begin{table*}[t!]
\centering
\scalebox{0.7}{
\begin{tabular}{|cc|ccccc|ccccc|ccccc|}
\hline
\multicolumn{2}{|c|}{\multirow{2}{*}{AP@all (\%)}}&\multicolumn{5}{c|}{BLESS}&\multicolumn{5}{c|}{EVALution}&\multicolumn{5}{c|}{Lenci/Benotto}\\ \cline{3-17}
& &CDE& $AL_1$& $\Delta$S& W$\cdot \Delta$S& C$\cdot \Delta$S& CDE& $AL_1$& $\Delta$S& W$\cdot\Delta$S& C$\cdot \Delta$S& CDE& $AL_1$& $\Delta$S& W$\cdot\Delta$S& C$\cdot \Delta$S \\  \hline %\cline{1-2}
\multicolumn{1}{|c|}{\multirow{4}{*}{SBOW}} & Freq&6.3&7.3&5.6&11.0&5.9&35.3&\textbf{32.6}&36.2&\textbf{33.0}&36.3&51.8&\textbf{47.6}&51.0&51.8&51.1\\
\multicolumn{1}{|c|}{}& PPMI&\textbf{13.6}&5.1&5.6&17.2&15.3&30.4&27.7&34.1&31.9&34.3&47.2&39.7&50.8&51.1&\textbf{52.0}\\
\multicolumn{1}{|c|}{}&PPMI w/  IS&6.2&5.0&5.5&12.4&5.8&\textbf{36.0}&27.5&\textbf{36.3}&32.9&\textbf{36.4}&\textbf{52.0}&43.1&50.9&\textbf{51.9}&50.7\\  
\multicolumn{1}{|c|}{}&All wiki&12.1&5.2&6.9&12.5&13.4&28.5&27.1&30.3&29.9&31.0&47.1&39.9&48.5&48.7&51.1\\  \cline{1-2}
\multicolumn{1}{|c|}{\multirow{3}{*}{DIVE}} &Full&9.3&\textbf{7.6}&6.0&\textbf{18.6}&\textbf{16.3}&30.0&27.5&34.9&32.3&33.0&46.7&43.2&\textbf{51.3}&51.5&50.4\\
\multicolumn{1}{|c|}{}& w/o PMI&7.8&6.9&5.6&16.7&7.1&32.8&32.2&35.7&32.5&35.4&47.6&44.9&50.9&51.6&49.7\\
\multicolumn{1}{|c|}{}& w/o IS&9.0&6.2&\textbf{7.3}&6.2&7.3&24.3&25.0&22.9&23.5&23.9&38.8&38.1&38.2&38.2&38.4\\  \cline{1-2}
\multicolumn{2}{|c|}{Kmean (Freq NMF)}&6.5&7.3&5.6&10.9&5.8&33.7&27.2&36.2&\textbf{33.0}&36.2&49.6&42.5&51.0&51.8&51.2\\ \hline
\hline
\multicolumn{2}{|c|}{\multirow{2}{*}{AP@all (\%)}}&\multicolumn{5}{c|}{Weeds}&\multicolumn{5}{c|}{Micro Average (4 datasets)}&\multicolumn{5}{c|}{Medical} \\  \cline{3-17}
& &CDE& $AL_1$& $\Delta$S& W$\cdot \Delta$S& C$\cdot \Delta$S& CDE& $AL_1$& $\Delta$S& W$\cdot\Delta$S& C$\cdot \Delta$S& CDE& $AL_1$& $\Delta$S& W$\cdot\Delta$S& C$\cdot \Delta$S \\  \hline 
\multicolumn{1}{|c|}{\multirow{4}{*}{SBOW}} & Freq &\textbf{69.5}&\textbf{58.0}&68.8&68.2&68.4&23.1&\textbf{21.8}&\textbf{22.9}&25.0&23.0 &19.4&\textbf{19.2}&\textbf{14.1}&18.4&15.3\\
\multicolumn{1}{|c|}{}& PPMI&61.0&50.3&\textbf{70.3}&\textbf{69.2}&69.3&\textbf{24.7}&17.9&22.3&28.1&\textbf{27.8}&\textbf{23.4}&8.7&13.2&20.1&\textbf{24.4}\\
\multicolumn{1}{|c|}{}&PPMI w/ IS&67.6&	52.2	&69.4	&68.7&	67.7&	23.2	&18.2	&\textbf{22.9}	&25.8	&22.9&22.8	&10.6	&13.7	&18.6&	17.0\\ 
\multicolumn{1}{|c|}{}&All wiki&61.3&48.6&70.0&68.5&\textbf{70.4}&23.4&17.7&21.7&24.6&25.8&22.3&	8.9&	12.2&	17.6&	21.1\\ \cline{1-2}
\multicolumn{1}{|c|}{\multirow{3}{*}{DIVE}} &Full &59.2&55.0&69.7&68.6&65.5&22.1&19.8&22.8&\textbf{28.9}&27.6&11.7&9.3&13.7&\textbf{21.4}&19.2\\
\multicolumn{1}{|c|}{}& w/o PMI&60.4&56.4&69.3&68.6&64.8&22.2&21.0&22.7&28.0&23.1&10.7 &8.4 &13.3&19.8&16.2\\
\multicolumn{1}{|c|}{}& w/o IS&49.2&47.3&45.1&45.1&44.9&18.9&17.3&17.2&16.8&17.5&10.9	&9.8	&7.4	&7.6&	7.7\\   \cline{1-2}
\multicolumn{2}{|c|}{Kmean (Freq NMF)}&69.4&51.1&68.8&68.2&68.9&22.5&19.3&\textbf{22.9}&24.9&23.0&12.6	&10.9	&14.0	&18.1&	14.6\\ \hline
\hline
\multicolumn{2}{|c|}{\multirow{2}{*}{AP@all (\%)}}&\multicolumn{5}{c|}{LEDS}&\multicolumn{5}{c|}{TM14}&\multicolumn{5}{c|}{Kotlerman 2010}\\ \cline{3-17}
&&CDE& $AL_1$& $\Delta$S& W$\cdot\Delta$S& C$\cdot \Delta$S&CDE& $AL_1$& $\Delta$S& W$\cdot\Delta$S& C$\cdot \Delta$S&CDE& $AL_1$& $\Delta$S& W$\cdot\Delta$S& C$\cdot \Delta$S \\  \hline
\multicolumn{1}{|c|}{\multirow{4}{*}{SBOW}} & Freq&82.7&70.4&70.7&83.3&73.3&55.6&53.2&54.9&55.7&55.0&35.9&\textbf{40.5}&\textbf{34.5}&37.0&35.4\\
\multicolumn{1}{|c|}{}& PPMI&\textbf{84.4}&50.2&72.2&\textbf{86.5}&\textbf{84.5}&56.2&52.3&54.4&57.0&\textbf{57.6}&\textbf{39.1}&30.9&33.0&37.0&36.3\\
\multicolumn{1}{|c|}{}&PPMI w/  IS&81.6&	54.5&	71.0&	84.7&	73.1&	\textbf{57.1}&	51.5&	55.1&	56.2&	55.4&	37.4&	31.0&	34.4&	37.8&	35.9\\  
\multicolumn{1}{|c|}{}&All wiki&83.1&	49.7&	67.9&	82.9&	81.4&	54.7&	50.5&	52.6&	55.1&	54.9&	38.5&	31.2&	32.2&	35.4	&35.3\\  \cline{1-2}
\multicolumn{1}{|c|}{\multirow{3}{*}{DIVE}} &Full&83.3&74.7&\textbf{72.7}&86.4&83.5&55.3&52.6&\textbf{55.2}&\textbf{57.3}&57.2&35.3&31.6&33.6&37.4&36.6\\
\multicolumn{1}{|c|}{}& w/o PMI&79.3&	\textbf{74.8}&	72.0&	85.5&	78.7&	54.7&	\textbf{53.9}&	54.9&	56.5&	55.4&	35.4&	38.9&	33.8&	\textbf{37.8}&	\textbf{36.7}\\
\multicolumn{1}{|c|}{}& w/o IS&64.6&	55.4&	43.2&	44.3&	46.1&	51.9&	51.2&	50.4&	52.0&	51.8&	32.9&	33.4&	28.1&	30.2&	29.7\\  \cline{1-2}
\multicolumn{2}{|c|}{Kmean (Freq NMF)}&80.3&	64.5&	70.7&	83.0&	73.0&	54.8&	49.0&	54.8&	55.6&	54.8&	32.1&	37.0&	34.5&	36.9&	34.8\\ \hline
%\multicolumn{2}{|c|}{SBOW Freq}&82.7&70.4&70.7&83.3&73.3&55.6&\textbf{53.2}&54.9&55.7&55.0&35.9&\textbf{40.5}&\textbf{34.5}&37.0&35.4 \\
%\multicolumn{2}{|c|}{SBOW PPMI}&\textbf{84.4}&50.2&72.2&\textbf{86.5}&\textbf{84.5}&\textbf{56.2}&52.3&54.4&57.0&\textbf{57.6}&\textbf{39.1}&30.9&33.0&37.0&36.3 \\
%\multicolumn{2}{|c|}{All wiki}&83.1&	49.7&	67.9&	82.9&	81.4&	54.7&	50.5&	52.6&	55.1&	54.9&	38.5&	31.2&	32.2&	35.4	&35.3 \\
%\multicolumn{2}{|c|}{DIVE}&83.3&\textbf{74.7}&\textbf{72.7}&86.4&83.5&55.3&52.6&\textbf{55.2}&\textbf{57.3}&57.2&35.3&31.6&33.6&\textbf{37.4}&\textbf{36.6} \\ \hline 
\hline \clineB{13-17}{2} 
\multicolumn{2}{|c|}{\multirow{2}{*}{AP@all (\%)}}&\multicolumn{5}{c|}{HypeNet}&\multicolumn{5}{c V{3}}{WordNet}&\multicolumn{5}{c V{3}}{\textbf{Micro Average (10 datasets)}}\\ \cline{3-17}
& & CDE& $AL_1$& $\Delta$S& W$\cdot \Delta$S& C$\cdot \Delta$S& CDE& $AL_1$& $\Delta$S& W$\cdot\Delta$S& \multicolumn{1}{c V{3}}{C$\cdot \Delta$S} & CDE& $AL_1$& $\Delta$S& W$\cdot\Delta$S& \multicolumn{1}{c V{3}}{C$\cdot \Delta$S} \\  \hline 
\multicolumn{1}{|c|}{\multirow{4}{*}{SBOW}} & Freq&37.5&\textbf{28.3}&46.9&\textbf{35.9}&43.4&56.6&55.2&55.5&56.2&\multicolumn{1}{c V{3}}{55.6}&31.1&\textbf{28.2}&31.5&31.6&\multicolumn{1}{c V{3}}{31.2}\\
\multicolumn{1}{|c|}{}& PPMI&23.8&24.0&47.0&32.5&33.1&57.7&53.9&55.6&56.8&\multicolumn{1}{c V{3}}{57.2}&30.1&23.0&31.1&32.9&\multicolumn{1}{c V{3}}{\textbf{33.5}}\\
\multicolumn{1}{|c|}{}&PPMI w/  IS&\textbf{38.5}&	26.7&	47.2&	35.5&	37.6&	57.0&	54.1&	55.7&	56.6&	\multicolumn{1}{c V{3}}{55.7}&	\textbf{31.8}&	24.1&	31.5&	32.1&	\multicolumn{1}{c V{3}}{30.3}\\  
\multicolumn{1}{|c|}{}&All wiki&23.0	&24.5&	40.5&	30.5&	29.7&	57.4&	53.1&	56.0&	56.4&	\multicolumn{1}{c V{3}}{57.3}&	29.0&	23.1&	29.2&	30.2&	\multicolumn{1}{c V{3}}{31.1}\\  \cline{1-2}
\multicolumn{1}{|c|}{\multirow{3}{*}{DIVE}} &Full&25.3&24.2&\textbf{49.3}&33.6&32.0&60.2&58.9&\textbf{58.4}&\textbf{61.1}&\multicolumn{1}{c V{3}}{\textbf{60.9}}&27.6&25.3&\textbf{32.1}&\textbf{34.1}&\multicolumn{1}{c V{3}}{32.7}\\
\multicolumn{1}{|c|}{}& w/o PMI&31.3&	27.0&	46.9&	33.8&	34.0&	59.2&	60.1&	58.2&	\textbf{61.1}&	\multicolumn{1}{c V{3}}{59.1}&	28.5&	26.7&	31.5&	33.4&	\multicolumn{1}{c V{3}}{30.1}\\
\multicolumn{1}{|c|}{}& w/o IS&20.1&	21.7&	20.3&	21.8&	22.0&	\textbf{61.0}&	56.3&	51.3&	55.7&	\multicolumn{1}{c V{3}}{54.7}&	22.3&	20.7&	19.1&	19.6&	\multicolumn{1}{c V{3}}{19.9}\\  \cline{1-2}
\multicolumn{2}{|c|}{Kmean (Freq NMF)}&33.7&	22.0&	46.0&	35.6&	\textbf{45.2}&	58.4&	\textbf{60.2}&	57.7&	60.1&	\multicolumn{1}{c V{3}}{57.9}&	29.1&	24.7&	31.5&	31.8&	\multicolumn{1}{c V{3}}{31.5}\\ \hline \clineB{13-17}{2}
%\multicolumn{2}{|c|}{SBOW Freq}&\textbf{37.5}&\textbf{28.3}&46.9&\textbf{35.9}&\textbf{43.4}&56.6&55.2&55.5&56.2&\multicolumn{1}{c V{3}}{55.6}&\textbf{31.1}&\textbf{28.2}&31.5&31.6&\multicolumn{1}{c V{3}}{31.2} \\
%\multicolumn{2}{|c|}{SBOW PPMI}&23.8&24.0&47.0&32.5&33.1&57.7&53.9&55.6&56.8&\multicolumn{1}{c V{3}}{57.2}&30.1&23.0&31.1&32.9&\multicolumn{1}{c V{3}}{\textbf{33.5}} \\
%\multicolumn{2}{|c|}{All wiki}&23.0	&24.5&	40.5&	30.5&	29.7&	57.4&	53.1&	56.0&	56.4&	\multicolumn{1}{c V{3}}{57.3}&	29.0&	23.1&	29.2&	30.2&	\multicolumn{1}{c V{3}}{31.1}\\
%\multicolumn{2}{|c|}{DIVE}&25.3&24.2&\textbf{49.3}&33.6&32.0&\textbf{60.2}&\textbf{58.9}&\textbf{58.4}&\textbf{61.1}&\multicolumn{1}{c V{3}}{\textbf{60.9}}&27.6&25.3&\textbf{32.1}&\textbf{34.1}&\multicolumn{1}{c V{3}}{32.7} \\ \hline \clineB{13-17}{2}
\end{tabular}
}
\caption{AP@all (\%) of 10 datasets. The box at lower right corner compares the micro average AP across all 10 datasets. Numbers in different rows come from different feature or embedding spaces. Numbers in different columns come from different datasets and unsupervised scoring functions. We also present the micro average AP across the first 4 datasets (BLESS, EVALution, Lenci/Benotto and Weeds), which are used as a benchmark for unsupervised hypernym detection~\citep{SantusSS17}. IS refers to inclusion shift on the shifted PMI matrix.}
\label{tb:AP_10}

\end{table*}

\begin{table*}[!t]
\begin{minipage}{0.5\textwidth}
\centering
\scalebox{0.7}{
\begin{tabular}{|cc|ccccc|}
\hline
\multicolumn{2}{|c|}{\multirow{2}{*}{Spearman $\rho$ (\%)}}&\multicolumn{5}{c|}{HyperLex} \\ \cline{3-7}
&&CDE& $AL_1$& $\Delta$S& W$\cdot\Delta$S& C$\cdot \Delta$S \\ \hline
\multicolumn{1}{|c|}{\multirow{4}{*}{SBOW}} & Freq&31.7&19.6&27.6&29.6&27.3\\
\multicolumn{1}{|c|}{}& PPMI&28.1&-2.3&\textbf{31.8}&\textbf{34.3}&\textbf{34.5}\\
\multicolumn{1}{|c|}{}&PPMI w/ IS&\textbf{32.4}&	2.1&	28.5&	31.0&	27.4\\  
\multicolumn{1}{|c|}{}&All wiki&25.3 & -2.2&28.0&30.5&	31.0\\  \cline{1-2}
\multicolumn{1}{|c|}{\multirow{3}{*}{DIVE}} &Full&28.9&18.7&31.2&33.3&32.8\\
\multicolumn{1}{|c|}{}& w/o PMI&29.2&	\textbf{22.2}&	29.5&	31.9&	29.2\\
\multicolumn{1}{|c|}{}& w/o IS&11.5&	-0.9&	-6.2&	-10.0&	-11.6\\  \cline{1-2}
\multicolumn{2}{|c|}{Kmean (Freq NMF)}&30.6&	3.3&	27.5&	29.5&	27.6\\ \hline
%SBOW Freq&\textbf{31.7}&\textbf{19.6}&27.6&29.6&27.3 \\
%SBOW PPMI&28.1&-2.3&\textbf{31.8}&\textbf{34.3}&\textbf{34.5} \\
%All wiki & 25.3 & -2.2&28.0&30.5&	31.0 \\
%DIVE&28.9&18.7&31.2&33.3&32.8 \\ \hline
\end{tabular}
}
\caption{Spearman $\rho$ (\%) in HyperLex.}
\label{tb:hyperlex}
\end{minipage}%
\begin{minipage}{0.5\textwidth}
\centering
\begin{tabular}{|c|c|c|}
\hline
SBOW Freq & SBOW PPMI & DIVE \\ \hline
5799& 3808&\textbf{20} \\ \hline
\end{tabular}
\caption{The average number of non-zero dimensions across all testing words in 10 datasets.}
\label{tb:avg_dim}
\end{minipage}
\end{table*}

\subsection{Baselines}
%Table~\ref{tb:baselines}
\begin{itemize}
\itemsep0em 
\item SBOW Freq: A word is represented by the frequency of its neighboring words. Applying PMI filter (set context feature to be $0$ if its value is lower than $\log(k_f)$) to SBOW Freq only makes its performances closer to (but still much worse than) SBOW PPMI, so we omit the baseline.
\item SBOW PPMI: SBOW which uses PPMI of its neighboring words as the features~\citep{bullinaria2007extracting}. Applying PMI filter to SBOW PPMI usually makes the performances worse, especially when $k_f$ is large. Similarly, a constant $\log(k')$ shifting to SBOW PPMI (i.e. $\max(PMI-\log(k'),0)$) is not helpful, so we set both $k_f$ and $k'$ to be 1.
\item SBOW PPMI w/ IS (with additional inclusion shift): The matrix reconstructed by DIVE when $k_I=1$. Specifically, $\textbf{w}[c]= \max(\log( \frac{P(w,c)}{P(w)*P(c)*\frac{Z}{\#(w)}}) ,0)$. 
\item SBOW all wiki: SBOW using PPMI features trained on the whole WaCkypedia.
\item DIVE without the PMI filter (DIVE w/o PMI) 
\item NMF on shifted PMI: Non-negative matrix factorization (NMF) on the shifted PMI without inclusion shift for DIVE (DIVE w/o IS). This is the same as applying the non-negative constraint on the skip-gram model. 
\item K-means (Freq NMF): The method first uses Mini-batch k-means~\citep{sculley2010web} to cluster words in skip-gram embedding space into 100 topics, and hashes each frequency count in SBOW into the corresponding topic. If running k-means on skip-grams is viewed as an approximation of clustering the SBOW context vectors, the method can be viewed as a kind of NMF~\citep{ding2005equivalence}.

\end{itemize}

DIVE performs non-negative matrix factorization on PMI matrix after applying inclusion shift and PMI filtering. To demonstrate the effectiveness of each step, we show the performances of DIVE after removing PMI filtering (DIVE w/o PMI), removing inclusion shift (DIVE w/o IS), and removing matrix factorization (SBOW PPMI w/ IS, SBOW PPMI, and SBOW all wiki). The methods based on frequency matrix are also tested (SBOW Freq and Freq NMF).

%(significantly decreases inclusion scoring functions and slightly increase generality measurement)
%apply PMI filter on SBOW PPMI (set context feature to be $0$ if the value is lower than $\log(k_f)$). We found that the performance degrades as $k_f$ increases. 

\subsection{Results and Discussions}
%In Table~\ref{tb:AP_10}, we compare different feature spaces using CDE, $AL_1$, summation difference ($\Delta$S), cosine similarity on skip gram times summation difference (W$\cdot\Delta$S), and cosine similarity on embeddings times summation difference (C$\cdot\Delta$S). 
%This is also why we evaluate our method on many datasets to make sure our conclusions hold in general. 
In Table~\ref{tb:AP_10}, we first confirm the finding of the previous review study of~\citet{SantusSS17}: there is no single hypernymy scoring function which always outperforms others. One of the main reasons is that different datasets collect negative samples differently. For example, if negative samples come from random word pairs (e.g. WordNet dataset), a symmetric similarity measure is a good scoring function. On the other hand, negative samples come from related or similar words in HypeNet, EVALution, Lenci/Benotto, and Weeds, so only estimating generality difference leads to the best (or close to the best) performance. The negative samples in many datasets are composed of both random samples and similar words (such as BLESS), so the combination of similarity and generality difference yields the most stable results.

DIVE performs similar or better on most of the scoring functions compared with SBOW consistently across all datasets in Table~\ref{tb:AP_10} and Table~\ref{tb:hyperlex}, while using many fewer dimensions (see Table~\ref{tb:avg_dim}). This leads to 2-3 order of magnitude savings on both memory consumption and testing time. Furthermore, the low dimensional embedding makes the computational complexity independent of the vocabulary size, which drastically boosts the scalability of unsupervised hypernym detection especially with the help of GPU. It is surprising that we can achieve such aggressive compression while preserving the similarity, generality, and inclusion signal in various datasets with different types of negative samples. Its results on C$\cdot \Delta$S and W$\cdot \Delta$S outperform SBOW Freq. Meanwhile, its results on $AL_1$ outperform SBOW PPMI. The fact that W$\cdot\Delta$S or C$\cdot\Delta$S usually outperform generality functions suggests that only memorizing general words is not sufficient. The best average performance on 4 and 10 datasets are both produced by W$\cdot\Delta $S on DIVE.

SBOW PPMI improves the W$\cdot\Delta$S and C$\cdot\Delta$S from SBOW Freq but sacrifices AP on the inclusion functions. It generally hurts performance to directly include inclusion shift in PPMI (PPMI w/ IS) or compute SBOW PPMI on the whole WaCkypedia (all wiki) instead of the first 51.2 million tokens. The similar trend can also be seen in Table~\ref{tb:hyperlex}. Note that $AL_1$ completely fails in the HyperLex dataset using SBOW PPMI, which suggests that PPMI might not necessarily preserve the distributional inclusion property, even though it can have good performance on scoring functions combining similarity and generality signals. 

Removing the PMI filter from DIVE slightly drops the overall precision while removing inclusion shift on shifted PMI (w/o IS) leads to poor performances. K-means (Freq NMF) produces similar AP compared with SBOW Freq but has worse $AL_1$ scores. Its best AP scores on different datasets are also significantly worse than the best AP of DIVE. This means that only making Word2Vec (skip-grams) non-negative or naively accumulating topic distribution in contexts cannot lead to satisfactory embeddings.

%the best metric in each dataset is usually combination metrics (i.e., W$\cdot\Delta$S or C$\cdot\Delta$S). The exceptions occur in Kotlerman 2010, HypeNet, where $AL_1$ and $\Delta$S are the best metrics, respectively. For WordNet, the best metric is word2vec (The performances of word2vec is shown in supplementary materials).

%The extensive experiments show that DIVE preserves the desirable properties of SBOW well. DIVE can even achieve slightly better performance than SBOW on some metrics. 

\section{Related Work}

Most previous unsupervised approaches focus on designing better hypernymy scoring functions for sparse bag of word (SBOW) features. They are well summarized in the recent study~\citep{SantusSS17}. \citet{SantusSS17} also evaluate the influence of different contexts, such as changing the window size of contexts or incorporating dependency parsing information, but neglect scalability issues inherent to SBOW methods. 

A notable exception is the Gaussian embedding model~\citep{VilnisM14}, which represents each word as a Gaussian distribution. However, since a Gaussian distribution is normalized, it is difficult to retain frequency information during the embedding process, and experiments on HyperLex~\citep{VulicGKHK16} demonstrate that a simple baseline only relying on word frequency can achieve good results. Follow-up work models contexts by a mixture of Gaussians~\citep{athiwaratkun2017multimodal} relaxing the unimodality assumption but achieves little improvement on hypernym detection tasks. 

\citet{kiela2015exploiting} show that images retrieved by a search engine can be a useful source of information to determine the generality of lexicons, but the resources (e.g. pre-trained image classifier for the words of interest) might not be available in many domains.

Order embedding~\citep{VendrovKFU15} is a supervised approach to encode many annotated hypernym pairs (e.g. all of the whole WordNet~\citep{miller1995wordnet}) into a compact embedding space, where the embedding of a hypernym should be smaller than the embedding of its hyponym in every dimension. Our method learns embedding from raw text, where a hypernym embedding should be larger than the embedding of its hyponym in every dimension. Thus, DIVE can be viewed as an unsupervised and reversed form of order embedding.

%Other semi-supervised hypernym detection methods aim to generalize from sets of annotated word pairs using raw text corpora. The goal of HyperVec~\citep{NguyenKWV17} is similar to our model: the embedding of a hypernym should be similar to its hyponym but with higher magnitude. However, their training process relies heavily on annotated hypernym pairs, and the performance drops significantly when reducing the amount of supervision. In addition to context distributions, previous work also leverages training data to discover useful text pattern indicating \smallt{is-a} relation~\citep{ShwartzGD16,RollerE16}, but it remains challenging to increase recall of hypernym detection because commonsense facts like \smallt{cat is-a animal} might not appear in the corpus.

Non-negative matrix factorization (NMF) has a long history in NLP, for example in the construction of topic models~\citep{pauca2004text}. Non-negative sparse embedding (NNSE)~\citep{murphy2012learning} and \citet{faruqui2015sparse} indicate that non-negativity can make embeddings more interpretable and improve word similarity evaluations. The sparse NMF is also shown to be effective in cross-lingual lexical entailment tasks but does not necessarily improve monolingual hypernymy detection~\citep{vyas2016sparse}. In our study, we show that performing NMF on PMI matrix with inclusion shift can preserve DIH in SBOW, and the comprehensive experimental analysis demonstrates its state-of-the-art performances on unsupervised hypernymy detection.

\section{Conclusions}
Although large SBOW vectors consistently show the best all-around performance in unsupervised hypernym detection, it is challenging to compress them into a compact representation which preserves inclusion, generality, and similarity signals for this task.
%Compressing unsupervised SBOW models into a compact representation is challenging while preserving inclusion, generality, and similarity signals which are important for hypernym detection. 
Our experiments suggest that the existing approaches and simple baselines such as Gaussian embedding, accumulating K-mean clusters, and non-negative skip-grams do not lead to satisfactory performance.

To achieve this goal, we propose an interpretable and scalable embedding method called \emph{distributional inclusion vector embedding (DIVE)} by performing non-negative matrix factorization (NMF) on a weighted PMI matrix. We demonstrate that scoring functions which measure inclusion and generality properties in SBOW can also be applied to DIVE to detect hypernymy, and DIVE performs the best on average, slightly better than SBOW while using many fewer dimensions. 

Our experiments also indicate that unsupervised scoring functions which combine similarity and generality measurements work the best in general, but no one scoring function dominates across all datasets. A combination of unsupervised DIVE with the proposed scoring functions produces new state-of-the-art performances on many datasets in the unsupervised regime. 
%is comparable with or even better than semi-supervised methods when the amount of training data are limited, and produces new state-of-the-art performances on many datasets under the unsupervised setup. 

\section{Acknowledgement}
This work was supported in part by the Center for Data Science and the Center for Intelligent Information Retrieval, in part by DARPA under agreement number FA8750-13-2-0020, in part by Defense Advanced Research Agency (DARPA) contract number HR0011-15-2-0036, in part by the National Science Foundation (NSF) grant numbers DMR-1534431 and IIS-1514053 and in part by the Chan Zuckerberg Initiative under the project “Scientific Knowledge Base Construction. The U.S. Government is authorized to reproduce and distribute reprints for Governmental purposes notwithstanding any copyright notation thereon. The views and conclusions contained herein are those of the authors and should not be interpreted as necessarily representing the official policies or endorsements, either expressed or implied, of DARPA, or the U.S. Government, or the other sponsors.

%\bibliography{naaclhlt2018}
\bibliography{ref}
\bibliographystyle{acl_natbib}

%\appendix

\newpage
%\clearpage

\section{Appendix}
\label{sec:supplemental}

In the appendix, we discuss our experimental details in Section~\ref{sec:exp_details}, the experiment for choosing representative scoring functions in Section~\ref{sec:exp2}, the performance comparison with previously reported results in Section~\ref{sec:reported_results}, the experiment of hypernym direction detection in Section~\ref{sec:gen_detect}, and an efficient way to computing $AL_1$ scoring function in Section~\ref{appendix:AL1}.

\subsection{Experimental details}
\label{sec:exp_details}

When performing the hypernym detection task, each paper uses different training and testing settings, and we are not aware of an standard setup in this field. For the setting which affects the performance significantly, we try to find possible explanations. For all the settings we tried, we do not find a setting choice which favors a particular embedding/feature space, and all methods use the same training and testing setup in our experiments.

\begin{table}[!t]
\centering
\scalebox{0.55}{
\begin{tabular}{|cc|cc|cc|cc|cc|}
\hline
\multicolumn{2}{|c|}{BLESS}&\multicolumn{2}{c|}{EVALution}&\multicolumn{2}{c|}{Lenci/Benotto}&\multicolumn{2}{c|}{Weeds}&\multicolumn{2}{c|}{Avg (4 datasets)} \\
N&OOV&N&OOV&N&OOV&N&OOV&N&OOV \\ 
26554&1507& 13675&2475& 5010& 1464&2928&643&48167&6089 \\ \hline
\multicolumn{2}{|c|}{Medical}&\multicolumn{2}{c|}{LEDS} & \multicolumn{2}{c|}{TM14}&\multicolumn{2}{c|}{Kotlerman 2010}&\multicolumn{2}{|c|}{HypeNet} \\
N&OOV&N&OOV&N&OOV&N&OOV&N&OOV \\
 12602&3711&2770&28 & 2188&178 &2940&89&17670&9424 \\ \hline
\multicolumn{2}{|c|}{WordNet} &\multicolumn{2}{c|}{Avg (10 datasets)}& \multicolumn{2}{c|}{HyperLex} \\
N&OOV &N&OOV &N&OOV \\
8000&3596 &94337&24110&2616&59 \\ \cline{1-6}
\end{tabular}
}
\caption{Dataset sizes. N denotes the number of word pairs in the dataset, and OOV shows how many word pairs are not processed by all the methods in Table~\ref{tb:AP_10} and Table~\ref{tb:hyperlex}.}
\label{tb:stats}

\end{table}

\begin{figure*}[t!]
\centering
\includegraphics[width=1\linewidth]{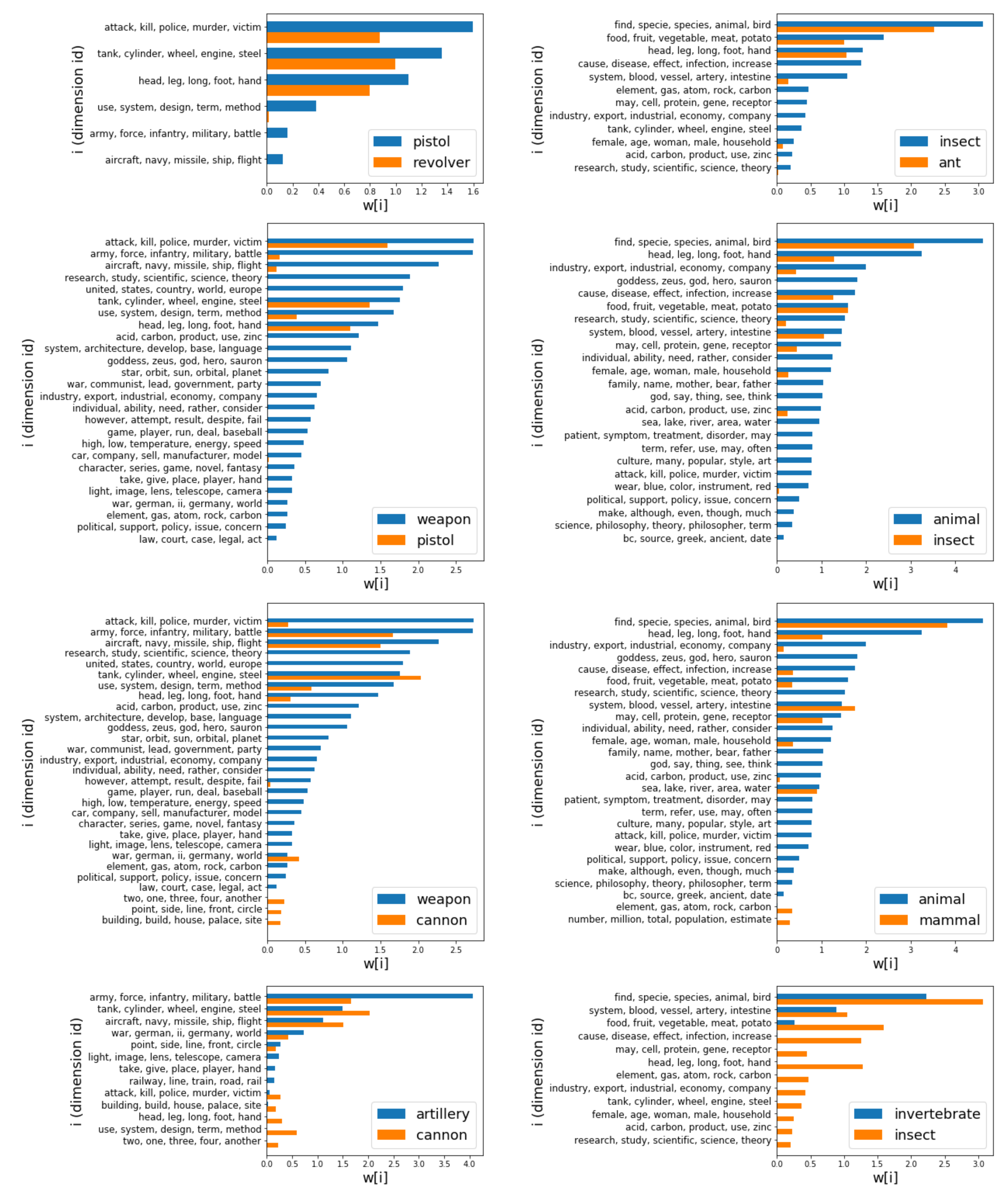}
\caption{Visualization of the DIVE embedding of word pairs with hypernym relation. The pairs include (\smallt{revolver},\smallt{pistol}), (\smallt{pistol},\smallt{weapon}), (\smallt{cannon},\smallt{weapon}), (\smallt{artillery},\smallt{cannon}), (\smallt{ant},\smallt{insect}), (\smallt{insect},\smallt{animal}), (\smallt{mammal},\smallt{animal}), and (\smallt{insect},\smallt{invertebrate}).}
\label{fig:hypernym_vis0}
\end{figure*}

\begin{figure*}[t!]
\centering
\includegraphics[width=1\linewidth]{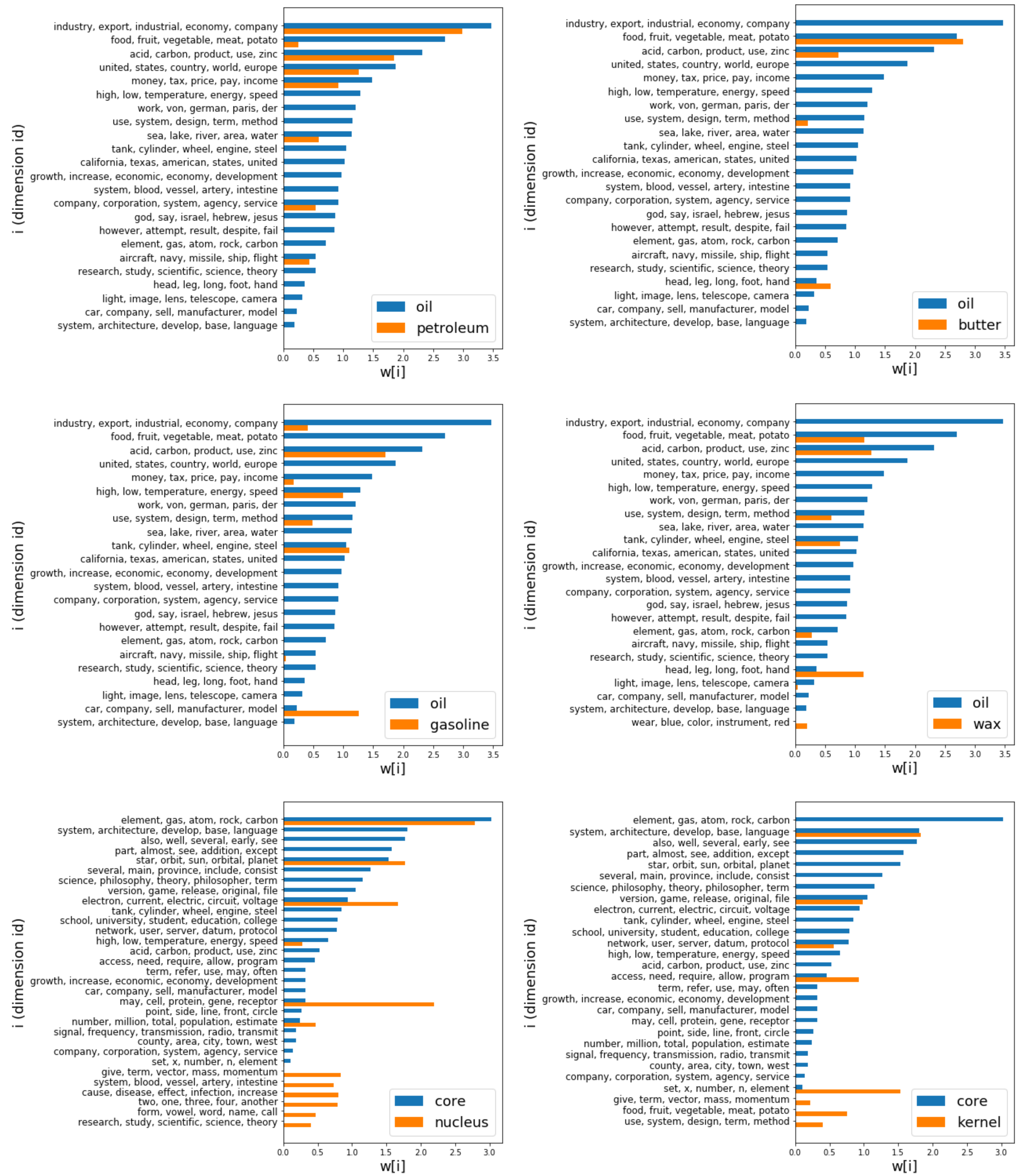}
\caption{Visualization of the DIVE embedding of \smallt{oil}, \smallt{core}, and their hyponyms.}
\label{fig:hypernym_vis}
\end{figure*}

\subsubsection{Training details}

We use WaCkypedia corpus~\citep{baroni2009wacky}, a 2009 Wikipedia dump, to compute SBOW and train the embedding. For the datasets without Part of Speech (POS) information (i.e. Medical, LEDS, TM14, Kotlerman 2010, and HypeNet), the training data of SBOW and embeddings are raw text. For other datasets, we concatenate each token with the Part of Speech (POS) of the token before training the models except the case when we need to match the training setup of another paper. All part-of-speech (POS) tags in the experiments come from NLTK. 

All words are lower cased. Stop words, rare words (occurs less than 10 times), and words including non-alphabetic characters are removed during our preprocessing step. To train embeddings more efficiently, we chunk the corpus into subsets/lines of 100 tokens instead of using sentence segmentation. Preliminary experiments show that this implementation simplification does not hurt the performance.
%(BLESS, EVALution, Lenci/Benotto, Weeds, WordNet, and HyperLex)

We train DIVE, SBOW, Gaussian embedding, and Word2Vec on only the first 512,000 lines (51.2 million tokens)\footnote{At the beginning, we train the model on this subset just to get the results faster. Later on, we find that in this subset of corpus, the context distribution of the words in testing datasets satisfy the DIH assumption better, so we choose to do all the comparison based on the subset.} because we find this way of training setting provides better performances (for both SBOW and DIVE) than training on the whole WaCkypedia or training on randomly sampled 512,000 lines. We suspect this is due to the corpus being sorted by the Wikipedia page titles, which makes some categorical words such as animal and mammal occur 3-4 times more frequently in the first 51.2 million tokens than the rest. The performances of training SBOW PPMI on the whole WaCkypedia is also provided for reference in Table~\ref{tb:AP_10} and Table~\ref{tb:hyperlex}. To demonstrate that the quality of DIVE is not very sensitive to the training corpus, we also train DIVE and SBOW PPMI  on PubMed and compare the performance of DIVE and SBOW PPMI on Medical dataset in Section~\ref{sec:pubmed}.

\subsubsection{Testing details}
The number of testing pairs N and the number of OOV word pairs is presented in Table~\ref{tb:stats}. 
The micro-average AP is computed by the AP of every datasets weighted by its number of testing pairs N.

In HypeNet and WordNet, some hypernym relations are determined between phrases instead of words. Phrase embeddings are composed by averaging embedding (DIVE, skip-gram), or SBOW features of each word. For WordNet, we assume the Part of Speech (POS) tags of the words are the same as the phrase. For Gaussian embedding, we use the average score of every pair of words in two phrases when determining the score between two phrases.

\subsubsection{Hyper-parameters}

For DIVE, the number of epochs is 15, the learning rate is 0.001, the batch size is 128, the threshold in PMI filter $k_f$ is set to be 30, and the ratio between negative and positive samples ($k_I$) is 1.5. The hyper-parameters of DIVE were decided based on the performance of HypeNet training set. The window size of skip-grams (Word2Vec) is 10. The number of negative samples ($k'$) in skip-gram is set as 5. 

For Gaussian embedding (GE), the number of mixture is 1, the number of dimension is 100, the learning rate is 0.01, the lowest variance is 0.1, the highest variance is 100, the highest Gaussian mean is 10, and other hyper-parameters are the default value in \url{https://github.com/benathi/word2gm}. The hyper-parameters of GE were also decided based on the performance of HypeNet training set. We also tried to directly tune the hyper-parameters on the micro-average performances of all datasets we are using (except HyperLex), but we found that the performances on most of the datasets are not significantly different from the one tuned by HypeNet.

\subsubsection{Kmeans as NMF}

For our K-means (Freq NMF) baseline, K-means hashing creates a $|V| \times 100$ matrix $G$ with orthonormal rows ($G^TG=I $), where $|V|$ is the size of vocabulary, and the $(i,k)$th element is $0$ if the word $i$ does not belong to cluster $k$.
Let the $|V| \times |V|$ context frequency matrix be denoted as $M_c$, where the $(i,j)$th element stores the count of word $j$ appearing beside word $i$. The $G$ created by K-means is also a solution of a type of NMF, where $M_c\approx FG^T$ and $G$ is constrained to be orthonormal~\citep{ding2005equivalence}. Hashing context vectors into topic vectors can be written as $M_cG\approx FG^TG = F$.

\subsection{Qualitative analysis}
\label{sec:qualitative}
To understand how DIVE preserves DIH more intuitively, we visualize the embedding of several hypernym pairs. In Figure~\ref{fig:hypernym_vis0}, we compare DIVE of different weapons and animals where the dimensions with the embedding value less than 0.1 are removed. We can see that hypernyms often have extra attributes/dimensions that their hyponyms lack. For example, \smallt{revolver} do not appears in the military context as often as \smallt{pistol} do and an \smallt{ant} usually does not cause diseases. We can also tell that \smallt{cannon} and \smallt{pistol} do not have hypernym relation because \smallt{cannon} appears more often in military contexts than \smallt{pistol}. 

In DIVE, the signal comes from the count of co-occurring context words. Based on DIH, we can know a terminology to be general only when it appears in diverse contexts many times. In Figure~\ref{fig:hypernym_vis0}, we illustrate the limitation of DIH by showing the DIVE of two relatively rare terminologies: \smallt{artillery} and \smallt{invertebrate}. There are other reasons that could invalid DIH. An example is that a specific term could appear in a special context more often than its hypernym~\citep{SantusSS17}. For instance, \smallt{gasoline} co-occurs with words related to cars more often than \smallt{oil} in Figure~\ref{fig:hypernym_vis}, and similarly for \smallt{wax} in contexts related to legs or foots. Another typical DIH violation is caused by multiple senses of words. For example, \smallt{nucleus} is the terminology for the core of atoms, cells, comets, and syllables. DIH is satisfied in some senses (e.g. the core of atoms) while not in other senses (the core of cells).

%For reference, Table~\ref{tb:baselines} shows AP@all using the following two baselines:
%\begin{itemize}
%\item Assigning a random score on each word pair (RND). 
%\item Cosine similarity using skip-grams (W). The window size of skip-grams (word2vec) is 10. The number of negative samples ($k'$) in skip-gram is set as 5. When composing skip gram into phrase embedding, average embedding is used.
%\end{itemize}

% \begin{table*}[!t]
% \centering
% \caption{AP@all (\%) for 10 datasets and Spearman $\rho$ (\%) for HyperLex. RND and W are random and Word2vec baselines, respectively. }
% \label{tb:baselines}
% \centering
% \scalebox{0.9}{
% \begin{tabular}{|cc|cc|cc|cc|cc|}
% \hline
% \multicolumn{2}{|c|}{BLESS}&\multicolumn{2}{c|}{EVALution}&\multicolumn{2}{c|}{Lenci/Benotto}&\multicolumn{2}{c|}{Weeds} &\multicolumn{2}{c|}{Avg (4 datasets)}\\
% RND&W&RND&W&RND&W&RND&W&RND&W \\
% 5.3& 9.2& 26.6& 25.4& 41.2& 40.8&51.4&51.6&17.9&19.7 \\ \hline
% \multicolumn{2}{|c|}{Medical}&\multicolumn{2}{c|}{LEDS} & \multicolumn{2}{c|}{TM14}&\multicolumn{2}{c|}{Kotlerman 2010}&\multicolumn{2}{c|}{HypeNet} \\
% RND&W&RND&W&RND&W &RND&W&RND&W\\
% 8.5&11.2&50.5&71.8&52.0&52.1 &30.8&39.5&24.5&20.7\\ \hline
% \multicolumn{2}{|c|}{WordNet} &\multicolumn{2}{c|}{Avg (10 datasets)}&  \multicolumn{2}{c|}{HyperLex}\\
% RND&W&RND&W&RND&W \\
% 55.2&63.0&23.2&25.3 &0&16.3\\ \cline{1-6}
% \end{tabular}
% }
% \end{table*}

\subsection{Hypernymy scoring functions analysis}
\label{sec:exp2}
Different scoring functions measure different signals in SBOW or embeddings. Since there are so many scoring functions and datasets available in the domain, we introduce and test the performances of various scoring functions so as to select the representative ones for a more comprehensive evaluation of DIVE on the hypernymy detection tasks. We denote the embedding/context vector of the hypernym candidate and the hyponym candidate as $\textbf{w}_p$ and $\textbf{w}_q$, respectively. 
%\citet{SantusSS17} have shown that there is no single scoring function which outperforms all others in all datasets.  
%, so an important research question is whether DIVE can preserve the desirable properties in SBOW which are measured by different scoring functions. 
%each scoring metric has its own advantage and
%perform differently on different datasets. 
%there are many other scoring metrics proposed. 

\subsubsection{Unsupervised scoring functions}

\begin{table*}[t!]
\centering
\scalebox{0.8}{
\begin{tabular}{cccccccc}
\hline
Word2Vec (W)  & Cosine (C)& SLQS Sub & SLQS Row ($\Delta$E) & Summation ($\Delta$S)	& Two norm ($\Delta$Q) \\
24.8 &26.7&27.4&27.6 & 31.5 & 31.2 \\ \hline
 W$\cdot\Delta$E	 & C$\cdot \Delta$E	 & W$\cdot \Delta$S& C$\cdot \Delta$S	& W$\cdot \Delta$Q& C$\cdot\Delta$Q \\ 
28.8&29.5	&\textbf{31.6}&31.2&31.4	&31.1 \\ \hline
 Weeds& CDE &	invCL	& Asymmetric L1 ($AL_1$)	\\
 19.0&31.1&30.7&28.2 \\ \hline
\end{tabular}
}
\caption{Micro average AP@all (\%) of 10 datasets using different scoring functions. The feature space is SBOW using word frequency.}
\label{tb:metrics}

\end{table*}

\textbf{Similarity}

A hypernym tends to be similar to its hyponym, so we measure the cosine similarity between word vectors of the SBOW features~\citep{levy2015supervised} or DIVE. We refer to the symmetric scoring function as Cosine or C for short in the following tables. We also train the original skip-grams with 100 dimensions and measure the cosine similarity between the resulting Word2Vec embeddings. This scoring function is referred to as Word2Vec or W.

\noindent \textbf{Generality}

The \emph{distributional informativeness hypothesis}~\citep{santus2014chasing} observes that in many corpora, semantically `general' words tend to appear more frequently and in more varied contexts. Thus, \citet{santus2014chasing} advocate using entropy of context distributions to capture the diversity of context. We adopt the two variations of the approach proposed by \citet{SantusSS17}: SLQS Row and SLQS Sub functions. We also refer to SLQS Row as $\Delta$E because it measures the entropy difference of context distributions. For SLQS Sub, the number of top context words is fixed as 100. 

Although effective at measuring diversity, the entropy totally ignores the frequency signal from the corpus. To leverage the information, we measure the generality of a word by its L1 norm ($||\textbf{w}_p||_1$) and L2 norm ($||\textbf{w}_p||_2$). Recall that Equation~\eqref{eq-reversed-order} indicates that the embedding of the hypernym $\textbf{y}$ should have a larger value at every dimension than the embedding of the hyponym $\textbf{x}$. When the inclusion property holds, $||\textbf{y}||_1 = \sum_i\textbf{y}[i] \geq \sum_i\textbf{x}[i] = ||\textbf{x}||_1$ and similarly $||\textbf{y}||_2 \geq ||\textbf{x}||_2$. Thus, we propose two scoring functions, difference of vector summation ($||\textbf{w}_p||_1-||\textbf{w}_q||_1$) and the difference of vector 2-norm ($||\textbf{w}_p||_2-||\textbf{w}_q||_2$). Notice that when applying the difference of vector summations (denoted as $\Delta$S) to SBOW Freq, it is equivalent to computing the word frequency difference between the hypernym candidate pair.

%we propose two simple approaches. 

\noindent \textbf{Similarity plus generality}

The combination of 2 similarity functions (Cosine and Word2Vec) and the 3 generality functions (difference of entropy, summation, and 2-norm of vectors) leads to six different scoring functions as shown in Table~\ref{tb:metrics}, and C$\cdot \Delta$S is the same scoring function we used in Experiment 1. It should be noted that if we use skip-grams with negative sampling (Word2Vec) as the similarity measurement (i.e., $W\cdot \Delta$ \{E,S,Q\}), the scores are determined by two embedding/feature spaces together (Word2Vec and DIVE/SBOW).

\begin{figure}[t!]
\centering
\includegraphics[width=0.9\linewidth]{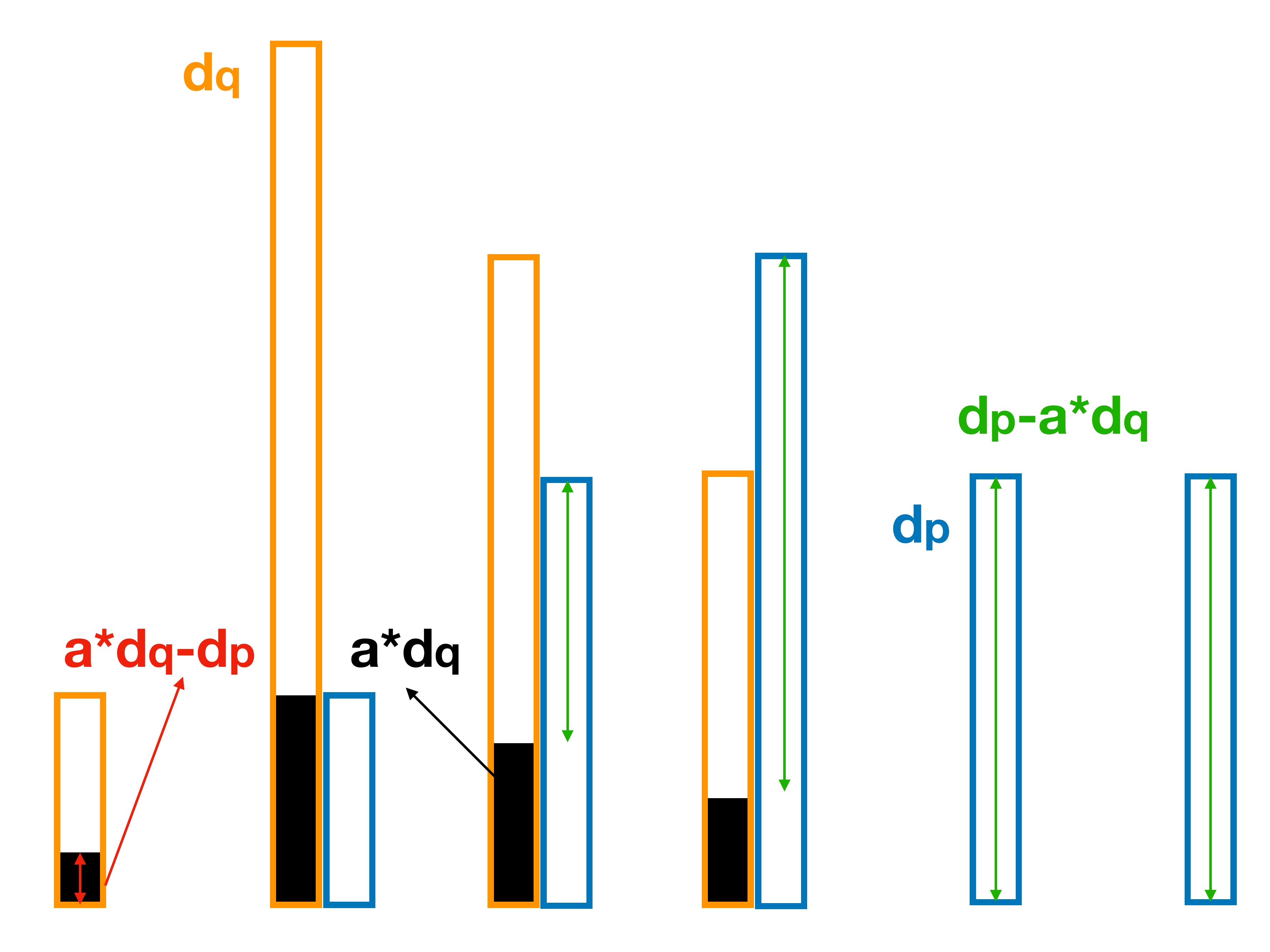}
\caption{An example of $AL_1$ distance. If the word pair indeed has the hypernym relation, the context distribution of hyponym ($\mathbf{d}_q$) tends to be included in the context distribution of hypernym ($\mathbf{d}_p$) after proper scaling according to DIH. Thus, the context words only appear beside the hyponym candidate ($a\mathbf{d}_q[c]-\mathbf{d}_p[c]$) causes higher penalty (weighted by $w_0$).}
\label{fig:AL1}
\end{figure}

\noindent \textbf{Inclusion}

Several scoring functions are proposed to measure inclusion properties of SBOW based on DIH. Weeds Precision~\citep{weeds2003general} and CDE~\citep{clarke2009context} both measure the magnitude of the intersection between feature vectors ($||\textbf{w}_p \cap \textbf{w}_q||_1$). For example, $\textbf{w}_p \cap \textbf{w}_q$ is defined by the element-wise minimum in CDE. Then, both scoring functions divide the intersection by the magnitude of the potential hyponym vector ($||\textbf{w}_q||_1$). invCL~\citep{lenci2012identifying} (A variant of CDE) is also tested. 

We choose these 3 functions because they have been shown to detect hypernymy well in a recent study~\citep{SantusSS17}. However, it is hard to confirm that their good performances come from the inclusion property between context distributions --- it is also possible that the context vectors of more general words have higher chance to overlap with all other words due to their high frequency. For instance, considering a one dimension feature which stores only the frequency of words, the naive embedding could still have reasonable performance on the CDE function, but the embedding in fact only memorizes the general words without modeling relations between words~\citep{levy2015supervised} and loses lots of inclusion signals in the word co-occurrence statistics.
%but the embedding loses lots of inclusion signals in the word co-occurrence statistics.
%~\citep{levy2015supervised}

In order to measure the inclusion property without the interference of the word frequency signal from the SBOW or embeddings, we propose a new measurement called asymmetric $L_1$ distance. We first get context distributions $\textbf{d}_p$ and $\textbf{d}_q$ by normalizing $\textbf{w}_p$ and $\textbf{w}_q$, respectively. Ideally, the context distribution of the hypernym $\textbf{d}_p$ will include $\textbf{d}_q$. This suggests the hypernym distribution $\textbf{d}_p$ is larger than context distribution of the hyponym with a proper scaling factor $a \textbf{d}_q$ (i.e., $\max(a \textbf{d}_q-\textbf{d}_p,0 )$ should be small). Furthermore,  both distributions should be similar, so $a \textbf{d}_q$ should not be too different from $\textbf{d}_p$ (i.e., $\max(\textbf{d}_p - a \textbf{d}_q,0 )$ should also be small). Therefore, we define asymmetric L1 distance as
\begin{equation}
\begin{aligned}
%\footnotesize
AL_1  = & \min_a \sum_c w_0 \cdot \max(a \textbf{d}_q[c]-\textbf{d}_p[c],0 )+ \\
& \max(\textbf{d}_p[c] - a \textbf{d}_q[c],0 ), 
\end{aligned} 
\end{equation}
where $w_0$ is a constant which emphasizes the inclusion penalty. If $w_0=1$ and $a=1$, $AL_1$ is equivalent to L1 distance. The lower $AL_1$ distance implies a higher chance of observing the hypernym relation. Figure~\ref{fig:AL1} illustrates a visualization of $AL_1$ distance. We tried $w_0=5$ and $w_0=20$. $w_0=20$ produces a worse micro-average AP@all on SBOW Freq, SBOW PPMI and DIVE, so we fix $w_0$ to be 5 in all experiments. An efficient way to solve the optimization in $AL_1$ is presented in Section~\ref{appendix:AL1}.

\subsubsection{Results and discussions}
We show the micro average AP@all on 10 datasets using different hypernymy scoring functions in Table~\ref{tb:metrics}. We can see the similarity plus generality signals such as C$\cdot\Delta$S and W$\cdot\Delta$S perform the best overall. Among the unnormalized inclusion based scoring functions, CDE works the best. $AL_1$ performs well compared with other functions which remove the frequency signal such as Word2Vec, Cosine, and SLQS Row. The summation is the most robust generality measurement. In the table, the scoring functions are applied to SBOW Freq, but the performances of hypernymy scoring functions on the other feature spaces (e.g. DIVE) have a similar trend.

\begin{table*}[t!]
\centering
\scalebox{0.8}{
\begin{tabular}{|c|c|c|c|c|c|c|c|}
\cline{1-6}
Dataset&BLESS&EVALution&LenciBenotto&Weeds&Medical \\ \cline{1-6}
Metric &\multicolumn{4}{c|}{AP@all }& F1 \\ \cline{1-6}
\multirow{2}{*}{Baselines} & \multicolumn{2}{c|}{invCL} & APSyn & CDE & Cosine \\ \cline{2-6}
& 5.1&\textbf{35.3}&38.2&44.1 & 23.1 \\ \cline{1-6}
DIVE + C$\cdot\Delta$S &  16.3 &33.0 &50.4 & 65.5 & 25.3 \\ \cline{1-6}
DIVE + W$\cdot\Delta$S & \textbf{18.6}  &32.3 &\textbf{51.5} & \textbf{68.6} & \textbf{25.7} \\ \hline
\hline
%\multicolumn{6}{|c|}{} \\ \hline
Dataset&LEDS&TM14&Kotlerman 2010&HypeNet & HyperLex \\ \hline
Metric&\multicolumn{3}{c|}{AP@all } &\multicolumn{1}{c|}{F1} &Spearman $\rho$ \\ \hline
\multirow{2}{*}{Baselines}  & \multicolumn{3}{c|}{balAPinc}  &	SLQS & Freq ratio \\ \cline{2-6}
 &73 &56&37 & 22.8 & 27.9 \\ \hline
DIVE + C$\cdot\Delta$S &  83.5 &57.2 & 36.6 &\textbf{41.9} & 32.8 \\ \hline
DIVE + W$\cdot\Delta$S & \textbf{86.4}  & \textbf{57.3} & \textbf{37.4} &38.6 & \textbf{33.3} \\ \hline

\end{tabular}
}
\caption{Comparison with previous methods based on sparse bag of word (SBOW). All values are percentages. The results of invCL~\citep{lenci2012identifying}, APSyn~\citep{santus2016unsupervised}, and CDE~\citep{clarke2009context} are selected because they have the best AP@100 in the first 4 datasets~\citep{SantusSS17}. Cosine similarity~\citep{levy2015supervised}, balAPinc~\citep{kotlerman2010directional} in 3 datasets~\citep{turney2015experiments}, SLQS~\citep{santus2014chasing} in HypeNet dataset~\citep{ShwartzGD16}, and Freq ratio (FR)~\citep{VulicGKHK16} are compared.}
\label{tb:SBOW_methods}
\end{table*}

\begin{table*}[t!]
\centering
\scalebox{0.8}{
\begin{tabular}{|c|c|c|c|c|c|c|}
\cline{1-6}
Dataset&HyperLex&EVALution&LenciBenotto&Weeds&Medical \\ \cline{1-6}
Metric& Spearman $\rho$ & \multicolumn{3}{c|}{AP@all}   &	\multicolumn{1}{c|}{F1}   \\ \cline{1-6}
Baselines &	\multicolumn{4}{c|}{HyperVec (1337)} & H-feature (897) \\ \cline{2-6}
(\#Training Hypernymy)  &30 & \textbf{39}	 &44.8 & 58.5  	&\textbf{26}  \\ \cline{1-6}
DIVE + C$\cdot\Delta$S (0) & \textbf{34.5} &   33.8&\textbf{52.9}&\textbf{70.0} &25.3 \\ \cline{1-6}
%DIVE + W$\cdot\Delta$S& \textbf{86.4} & 33.9&\textbf{54.5}&\textbf{73.5}&\textbf{35.7}&25.7 \\ \cline{1-6}
%\multicolumn{7}{|c|}{} \\ \cline{1-7}
\end{tabular}
}
\caption{Comparison with semi-supervised embeddings (with limited training data). All values are percentages. The number in parentheses beside each approach indicates the number of annotated hypernymy word pairs used to train the model. Semi-supervised embeddings include HyperVec~\citep{NguyenKWV17} and H-feature~\citep{RollerE16}. Note that HyperVec ignores POS in the testing data, so we follow the setup when comparing with it.}
\label{tb:embedding_methods}
\end{table*}

\subsection{Comparison with reported results}
\label{sec:reported_results}
Each paper uses slightly different setups\footnote{Notice that some papers report F1 instead of AP. When comparing with them, we use 20 fold cross validation to determine prediction thresholds, as done by~\citet{RollerE16}.}, so it is hard to conclude that our methods are better. However, by comparing with reported numbers, we would like to show that the unsupervised methods seem to be previously underestimated, and it is possible for the unsupervised embeddings to achieve performances which are comparable with semi-supervised embeddings when the amount of training data is limited. 

\subsubsection{Comparison with SBOW}
In Table~\ref{tb:SBOW_methods}, DIVE with two of the best scoring functions (C$\cdot\Delta$S and W$\cdot\Delta$S) is compared with the previous unsupervised state-of-the-art approaches based on SBOW on different datasets. 

There are several reasons which might cause the large performance gaps in some datasets. In addition to the effectiveness of DIVE, some improvements come from our proposed scoring functions. The fact that every paper uses a different training corpus also affects the performances. Furthermore, \citet{SantusSS17} select the scoring functions and feature space for the first 4 datasets based on AP@100, which we believe is too sensitive to the hyper-parameter settings of different methods.

%Previous work uses different scoring functions.
%It bears mentioning that the SBOW Freq and SBOW PPMI are also tested in the previous literature review~\citep{SantusSS17}. One of the possible reasons that DIVE can achieve much better results is because they select scoring functions and context types based on AP@100, which we believe is too sensitive to the hyper-parameter settings of different methods. Other reasons might lie in the effectiveness of newly proposed combination functions and different training corpora are used.

\subsubsection{Comparison with semi-supervised embeddings}

In addition to the unsupervised approach, we also compare DIVE with semi-supervised approaches. When there are sufficient training data, there is no doubt that the semi-supervised embedding approaches such as HypeNet~\citep{ShwartzGD16}, H-feature detector~\citep{RollerE16}, and HyperVec~\citep{NguyenKWV17} can achieve better performance than all unsupervised methods. However, in many domains such as scientific literature, there are often not many annotated hypernymy pairs (e.g. Medical dataset~\citep{LevyDG14}). 

Since we are comparing an unsupervised method with semi-supervised methods, it is hard to fairly control the experimental setups and tune the hyper-parameters. In Table~\ref{tb:embedding_methods}, we only show several performances which are copied from the original paper when training data are limited\footnote{We neglect the performances from models trained on more than 10,000 hypernym pairs, models trained on the same evaluation datasets with more than 1000 hypernym pairs using cross-validation, and models using other sources of information such as search engines and image classifiers (e.g. the model from \citet{kiela2015exploiting}).}. As we can see, the performance from DIVE is roughly comparable to the previous semi-supervised approaches trained on small amount of hypernym pairs. This demonstrates the robustness of our approach and the difficulty of generalizing hypernymy annotations with semi-supervised approaches.

%Notice that the AP of baselines is copied from the original paper, so .
%there must be several setup differences between paper, the results might not be directly comparable. 

\begin{table*}[t!]
\centering
\begin{tabular}{|c|cccccc|}
\hline
Query & \multicolumn{6}{c|}{Top 30 general words} \\ 
\hline
 &	use&	name&	system&	include&	base&	city \\
&	large&	state&	group&	power&	death&	form \\
&	american&	life&	may&	small&	find&	body \\
&	design&	work&	produce&	control&	great&	write \\
&	study&	lead&	type&	people&	high&	create \\
\hline
\multirow{5}{*}{species}&	specie&	species&	animal&	find&	plant&	may \\
&	human&	bird&	genus&	family&	organism&	suggest \\
&	gene&	tree&	name&	genetic&	study&	occur \\
&	fish&	disease&	live&	food&	cell&	mammal \\
&	evidence&	breed&	protein&	wild&	similar&	fossil \\
\hline
\multirow{5}{*}{system} & system&	use&	design&	provide&	operate&	model \\
 &	standard&	type&	computer&	application & develop&	method \\
 & allow&	function&	datum&	device&	control&	information \\	
 & process&	code& via&	base&	program&	software \\	
 & network & file&	development&	service&	transport&	law \\
\hline
\end{tabular}

\caption{We show the top 30 words with the highest embedding magnitude after dot product with the query embedding $\mathbf{q}$ (i.e. showing w such that $||\mathbf{w}^T\mathbf{q}||_1$ is one of the top 30 highest values). The rows with the empty query word sort words based on $||\mathbf{w}||_1$.}
\label{tb:gen_vis}

\end{table*}

\begin{table}[t!]
\centering
\scalebox{0.9}{
\begin{tabular}{|c|c|}
\hline
\multicolumn{2}{|c|}{Micro Average (10 datasets)} \\ \hline
SBOW Freq + SLQS Sub& SBOW Freq + $\triangle$S \\ \hline
 64.4 &66.8 \\ \hline
SBOW PPMI + $\triangle$S& DIVE + $\triangle$S \\ 
66.8&\textbf{67.0} \\ \hline
\end{tabular}
}
\caption{Accuracy (\%) of hypernym directionality prediction across 10 datasets.}
\label{tb:directionality}
\end{table}

\subsection{Generality estimation and hypernym directionality detection}
\label{sec:gen_detect}
In Table~\ref{tb:gen_vis}, we show the most general words in DIVE under different queries as constraints. We also present the accuracy of judging which word is a hypernym (more general) given word pairs with hypernym relations in Table~\ref{tb:directionality}. The direction is classified correctly if the generality score is greater than 0 (hypernym is indeed predicted as the more general word). For instance, summation difference ($\Delta$S) classifies correctly if $||\textbf{w}_p||_1-||\textbf{w}_q||_1>0$ ($||\textbf{w}_p||_1>||\textbf{w}_q||_1$). 

From the table, we can see that the simple summation difference performs better than SQLS Sub, and DIVE predicts directionality as well as SBOW. Notice that whenever we encounter OOV, the directionality is predicted randomly. If OOV is excluded, the accuracy of predicting directionality using unsupervised methods can reach around 0.7-0.75.

\subsection{PubMed experiment}
\label{sec:pubmed}
To demonstrate that DIVE can compress SBOW in a different training corpus, we train DIVE and SBOW PPMI on biomedical paper abstracts in a subset of PubMed~\citep{wei2012accelerating} and compare their performances on Medical dataset~\citep{LevyDG14}. We randomly shuffle the order of abstracts, remove the stop words, and only use the first 51.2 million tokens. The same hyper-parameters of DIVE and SBOW PPMI are used, and their AP@all are listed in Table~\ref{tb:PubMed}. For most scoring functions, the AP@all difference is within ~1\% compared with the model trained by WaCkypedia.

\begin{table}[t!]
\centering
\scalebox{0.7}{
\begin{tabular}{|cc|ccccc|}
\hline
\multicolumn{2}{|c|}{\multirow{2}{*}{AP@all (\%)}}&\multicolumn{5}{c|}{Medical} \\ \cline{3-7}
& &CDE& $AL_1$& $\Delta$S& W$\cdot \Delta$S& C$\cdot \Delta$S \\ \hline
\multicolumn{1}{|c|}{\multirow{2}{*}{SBOW PPMI}}&wiki&\textbf{23.4}&8.7&13.2&20.1&\textbf{24.4}\\
\multicolumn{1}{|c|}{} &PubMed & 20.0 & 7.2 & 14.2 & 21.1 & 23.5\\ \cline{1-2}
\multicolumn{1}{|c|}{\multirow{2}{*}{DIVE}}&wiki&11.7&\textbf{9.3}&13.7&\textbf{21.4}&19.2\\
\multicolumn{1}{|c|}{} &PubMed &12.6 & \textbf{9.3} & \textbf{15.9} & 21.2 & 20.4\\
\hline
\end{tabular}
}
\caption{Training corpora comparison}
\label{tb:PubMed}

\end{table}

\subsection{Efficient way to compute asymmetric L1 ($AL_1$)}
\label{appendix:AL1}
Recall that Equation~\eqref{eq:AL1} defines $AL_1$ as follows: 
\begin{align*}
\mathcal{L} = & \min_a~~ \sum_c w_0 \max(a\mathbf{d}_q[c]-\mathbf{d}_p[c],0)+ \\ & \max(\mathbf{d}_p[c]-a\mathbf{d}_q[c],0),
%AL_1=\mathcal{L} = \min_a~~ \sum_c w_0 & \max(a\mathbf{d}_q[c]-\mathbf{d}_p[c],0)+ \\
%& \max(\mathbf{d}_p[c]-a\mathbf{d}_q[c],0),
\end{align*}
where $\mathbf{d}_p[c]$ is one of dimension in the feature vector of hypernym $\mathbf{d}_p$, $a\mathbf{d}_q$ is the feature vector of hyponym after proper scaling. In Figure~\ref{fig:AL1}, an simple example is visualized to illustrate the intuition behind the distance function.

By adding slack variables $\zeta$ and $\xi$, the problem could be converted into a linear programming problem:
\begin{align*}
\mathcal{L} = \min_{a,\zeta,\xi}&~~ w_0\sum_c\zeta_c+\sum_c\xi_c\\
&\zeta_c \ge a\mathbf{d}_q[c]-\mathbf{d}_p[c],~~~\zeta_c\ge0\\
&\xi_c \ge \mathbf{d}_p[c]-a\mathbf{d}_q[c],~~~\xi_c\ge0\\
&a\ge0,
\end{align*}
so it can be simply solved by a general linear programming library. 

Nevertheless, the structure in the problem actually allows us to solve this optimization by a simple sorting. In this section, we are going to derive the efficient optimization algorithm. 

By introducing Lagrangian multiplier for the constraints, we can rewrite the problem as
\begin{align*}
\mathcal{L}= & \min_{a,\zeta,\xi}\max_{\alpha,\beta,\gamma,\delta}~~ w_0\sum_c\zeta_c+\sum_c\xi_c \\
& -\sum_c\alpha_c(\zeta_c-a\mathbf{d}_q[c]+\mathbf{d}_p[c]) \\ 
&-\sum_c\beta_c(\xi_c-\mathbf{d}_p[c]+a\mathbf{d}_q[c]) \\
& -\sum_c\gamma_c\zeta_c - \sum_c\delta_c\xi_c \\
& \zeta_c\ge0, \xi_c\ge0, \alpha_c\ge0, \beta_c\ge0, \\
& \gamma_c\ge0, \delta_c\ge0, a\ge0
\end{align*}

First, we eliminate the slack variables by taking derivatives with respect to them:
\begin{align*}
\frac{\partial\mathcal{L}}{\partial \zeta_c}=0&=1-\beta_c-\delta_c\\
\delta_c&=1-\beta_c,~~~\beta_c\le1\\
\frac{\partial\mathcal{L}}{\partial \xi_c}=0&=1-\gamma_c-\alpha_c\\
\gamma_c&=w_0-\alpha_c,~~~\alpha_c\le w_0.\\
\end{align*}

By substituting in these values for $\gamma_c$ and $\delta_c$, we get rid of the slack variables and have a new Lagrangian:
\begin{align*}
\mathcal{L}=\min_{a}\max_{\alpha,\beta}& -\sum_c\alpha_c(-a\mathbf{d}_q[c]+\mathbf{d}_p[c]) \\
& -\sum_c\beta_c(-\mathbf{d}_p[c]+a\mathbf{d}_q[c]) \\
&0\le\alpha_c\le w_0,0 \le \beta_c\le 1,  a \ge 0
\end{align*}

We can introduce a new dual variable $\lambda_c=\alpha_c-\beta_c+1$ and rewrite this as:
\begin{align*}
\mathcal{L}=\min_{a}\max_{\lambda}&~~ \sum_c(\lambda_c - 1)(a\mathbf{d}_q[c]-\mathbf{d}_p[c]) \\
&0\le\lambda_c\le w_0+1, a \ge 0
\end{align*}

Let's remove the constraint on $a$ and replace with a dual variable $\eta$:
\begin{align*}
\mathcal{L}=\min_{a}\max_{\lambda}&~~ \sum_c(\lambda_c - 1)(a\mathbf{d}_q[c]-\mathbf{d}_p[c]) -\eta a \\
&0\le\lambda_c\le w_0+1, \eta \ge 0
\end{align*}

Now let's differentiate with respect to $a$ to get rid of the primal objective and add a new constraint:
\begin{align*}
\frac{\partial\mathcal{L}}{\partial a}=0&=\sum_c\lambda_c\mathbf{d}_q[c]-\sum_c \mathbf{d}_q[c] -\eta\\
\sum_c\lambda_c \mathbf{d}_q[c]& =\sum_c \mathbf{d}_q[c]+\eta\\
\mathcal{L}=\max_{\lambda}&~~  \sum_c \mathbf{d}_p[c] - \sum_c\lambda_c \mathbf{d}_p[c]\\
\sum_c\lambda_c \mathbf{d}_q[c]& =\sum_c \mathbf{d}_q[c]+\eta\\
&0\le\lambda_c\le w_0+1, \eta \ge 0
\end{align*}

Now we have some constant terms that are just the sums of $\mathbf{d}_p$ and $\mathbf{d}_q$, which will be 1 if they are distributions.
\begin{align*}
\mathcal{L}=\max_{\lambda}&~~ 1 - \sum_c\lambda_c \mathbf{d}_p[c]\\
&\sum_c\lambda_c \mathbf{d}_q[c]=1+\eta\\
&0\le\lambda_c\le w_0+1, \eta \ge 0
\end{align*}

Now we introduce a new set of variables $\mu_c=\lambda_c \mathbf{d}_q[c]$ and we can rewrite the objective as:
\begin{align*}
\mathcal{L}=\max_{\mu}&~~ 1 - \sum_c\mu_c\frac{\mathbf{d}_p[c]}{\mathbf{d}_q[c]}\\
&\sum_c\mu_c=1+\eta\\
&0\le\mu_c\le (w_0+1)\mathbf{d}_q[c],\eta \ge 0
\end{align*}

Note that for terms where $\mathbf{d}_q[c]=0$ we can just set $\mathbf{d}_q[c]=\epsilon$ for some very small epsilon, and in practice, our algorithm will not encounter these because it sorts.

So $\mu$ we can think of as some fixed budget that we have to spend up until it adds up to 1, but it has a limit of how much we can spend for each coordinate, given by $(w_0+1)\mathbf{d}_q[c]$. Since we're trying to minimize the term involving $\mu$, we want to allocate as much budget as possible to the smallest terms in the summand, and then 0 to the rest once we've spent the budget. This also shows us that our optimal value for the dual variable $\eta$ is just 0 since we want to minimize the amount of budget we have to allocate.

To make presentation easier, lets assume we sort the vectors in order of increasing $\frac{\mathbf{d}_p[c]}{\mathbf{d}_q[c]}$, so that $\frac{\mathbf{d}_p[1]}{\mathbf{d}_q[1]}$ is the smallest element, etc. We can now give the following algorithm to find the optimal $\mu$.

\begin{align*}
&\text{init}~~S=0,c=1,\mu=0\\
&\text{while}~~S\le 1:\\
&~~~~\mu_c=\min(1-S,(w_0+1)\mathbf{d}_q[c])\\
&~~~~S=S+\mu_c\\
&~~~~c=c+1
\end{align*}

At the end we can just plug in this optimal $\mu$ to the objective to get the value of our scoring function.

\end{document}